\documentclass[lettersize,journal]{IEEEtran}
\usepackage{amsmath,amsfonts}
\usepackage{algorithmic}
\usepackage{algorithm}
\usepackage{array}
\usepackage[caption=false,font=normalsize,labelfont=sf,textfont=sf]{subfig}
\usepackage{textcomp}
\usepackage{stfloats}
\usepackage{url}
\usepackage{verbatim}
\usepackage{graphicx}
\usepackage{cite}
\usepackage{hyperref}
\usepackage{rotating}
\usepackage{adjustbox}
\usepackage{booktabs}
\usepackage{multirow}
\usepackage[table]{xcolor}

\begin{document}

\title{MetaEarth-MM: Unified Multimodal Remote Sensing Image Generation with Scene-centered Joint Modeling}

\author{Zhiping Yu, Chenyang Liu, Jinqi Cao, Qinzhe Yang, Siwei Yu,  Zhengxia Zou, \IEEEmembership{Senior Member,~IEEE}, and\\ Zhenwei Shi*, \IEEEmembership{Senior Member,~IEEE}

\thanks{This work was supported in part by the National Natural Science Foundation of China under Grants U24B20177, 62125102, U25A20401, 62471014 and 624B2017, in part by the Inner Mongolia Autonomous Region Science and Technology Planning Project under Grant 2025YFHH0124, in part by the Beijing Natural Science Foundation under QY25221 and in part by the Fundamental Research Funds for the Central Universities.  
\textit{(*Corresponding author: Zhenwei Shi.)}}
\thanks{Zhiping Yu, Chenyang Liu, Jinqi Cao, Siwei Yu, Zhengxia Zou and Zhenwei Shi are with the Department of Aerospace Intelligent Science and Technology, School of Astronautics, Beihang University, Beijing 100191, China and the State Key Laboratory of Virtual Reality Technology and Systems, Beihang University, Beijing 100191, China (e-mail:
shizhenwei@buaa.edu.cn).}
\thanks{Qinzhe Yang is with Shenyuan Honors College, Beihang University, Beijing 100191, China.}
}

\markboth{ }%
{Shell \MakeLowercase{\textit{et al.}}: A Sample Article Using IEEEtran.cls for IEEE Journals}


\maketitle

\begin{abstract}
Multi-modal remote sensing images are vital for Earth observation, yet complete paired observations are often scarce in practice. Existing generative methods commonly address this problem through isolated pairwise modality translation, but their versatility and scalability remain limited as the number of modalities and generation tasks increases. Here, we develop a generative foundation model MetaEarth-MM for multi-modal remote sensing imagery, enabling paired joint generation and any-to-any translation across five modalities within a unified model. Recognizing the intrinsic scene consistency underlying multi-modal observations, we introduce a scene-centered joint modeling paradigm in MetaEarth-MM. Unlike previous methods that rely on direct appearance-level cross-modal mapping, our model organizes the generation around the underlying scene content. Specifically, MetaEarth-MM adopts a decoupled architecture that first infers a latent scene representation from available observations, and then generates target modalities conditioned on this intermediate state. To support training, we further construct EarthMM, a large-scale dataset comprising 2.8 million multi-resolution global images with 2.2 million aligned pairs. Extensive experiments demonstrate that MetaEarth-MM not only exhibits strong generative capability and robust generalization across diverse generation tasks, but also supports downstream tasks at both data and representation levels, highlighting its potential as a general foundation model for cross-modal Earth observation. The code and dataset will be available at \url{https://github.com/YZPioneer/MetaEarth-MM}.

\end{abstract}


\section{Introduction}
\IEEEPARstart{M}{ulti-modal} imagery is fundamental to modern Earth observation, as different remote sensing modalities provide complementary information about the Earth's surface~\cite{ringmoe, crossearth,liu2025remote}. Modalities such as optical imagery, Synthetic Aperture Radar (SAR), and Near Infrared (NIR) capture distinct physical properties and play important roles in applications including environmental monitoring, urban planning, and disaster response~\cite{skysens, hypersigma,liu2024changeAgent,liu2024rscama}. However, complete and well-aligned paired observations across multiple modalities are often difficult to obtain in practice. This data scarcity has motivated growing interest in generative models for multi-modal remote sensing image synthesis~\cite{10752552, ph-gan, rsdiffusion_review}.

Most existing studies address this problem through modality translation~\cite{pix2pix, text2earth, earthmapper, dogan, hsigene}, where a model learns a conditional distribution from an observed source modality to a target modality. While this paradigm has achieved promising results for specific modality pairs, it remains limited as a general solution for multi-modal remote sensing generation. On the one hand, it inherently relies on the presence of an observed source modality, limiting its applicability to source-free paired generation. On the other hand, its scalability remains limited, as extending to more modalities and generation tasks often requires training separate translation models for different modality pairs. These limitations highlight the need to develop a unified generative model for diverse modalities and multi-modal generation tasks.

\begin{figure}[t]
    \centering
    \includegraphics[width=0.9\linewidth]{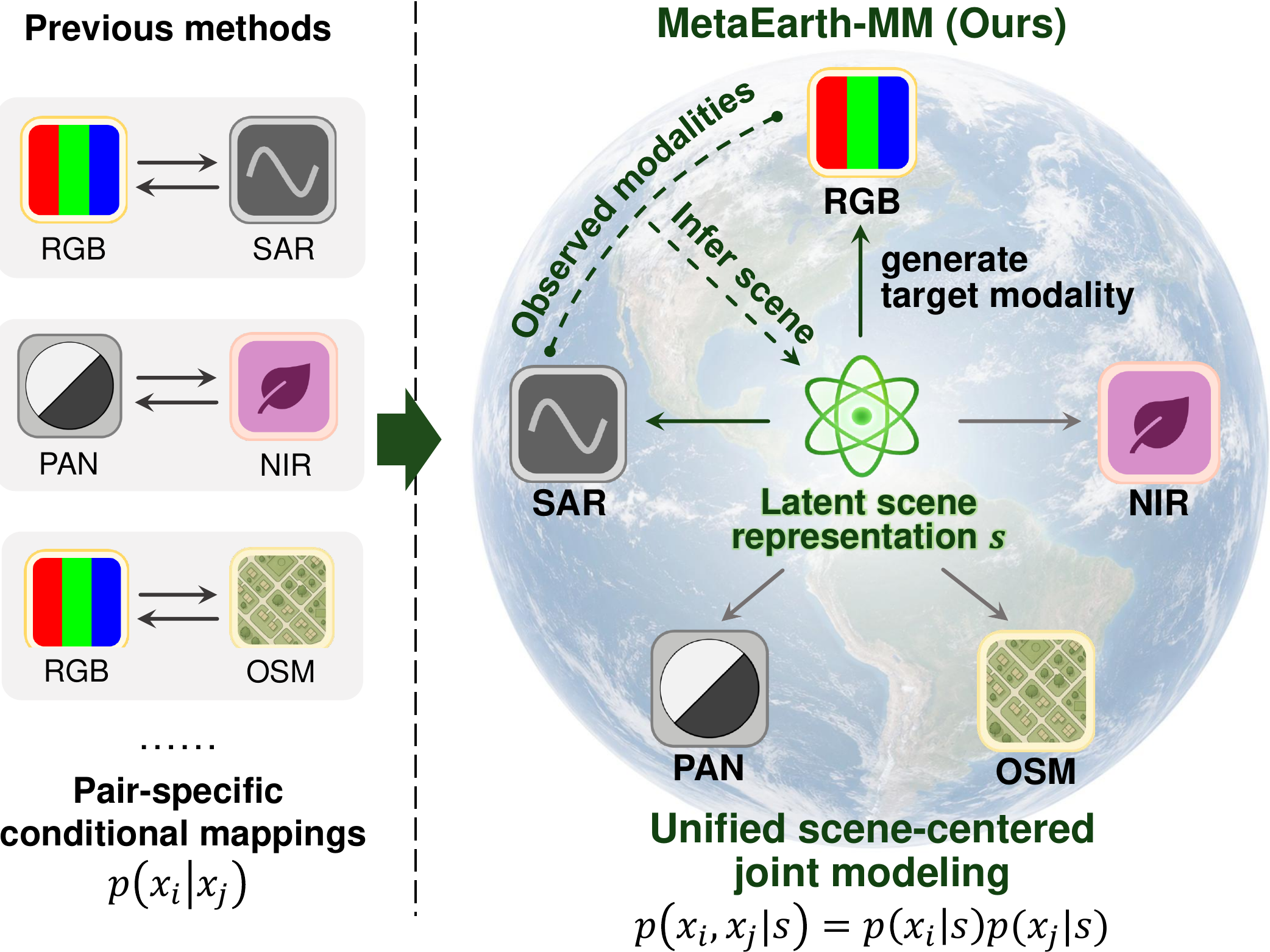}
    \caption{Comparison between existing pairwise modality translation methods and the proposed MetaEarth-MM. \textbf{Left}: Previous isolated direct cross-modal mappings. \textbf{Right}: Proposed MetaEarth-MM with scene-centered joint modeling. Unlike previous pair-specific appearance mappings, our model organizes multi-modal generation around an inferred latent scene representation.}
    \label{fig:teaser}
\end{figure}

Joint distribution modeling provides an effective way to construct such a unified model. Under this paradigm, pairwise conditional distributions are subsumed by joint modeling\cite{unidiffuser, omniflow}. The same model can therefore generate a target modality when an observed modality is available, and jointly sample paired modalities without relying on a specific source modality. Moreover, by capturing cross-modal dependencies and co-occurrence patterns, joint modeling further enables shared scene-level information and cross-modal complementarities to be integrated across different modality pairs. This provides a more general basis for building a unified multi-modal generative model.

Despite its promise, developing a unified multi-modal generative model remains highly challenging due to the heterogeneity across diverse modalities. Different modalities exhibit distinct visual characteristics, physical properties, and data distributions. For example, optical imagery mainly reflects spectral appearance such as color and texture, whereas SAR captures microwave backscatter and primarily characterizes structural and dielectric properties. These modality-specific patterns may interfere with one another in a unified model, leading to degraded generation fidelity. In addition, sharing a limited model capacity across different modalities will further complicate joint optimization within a unified network.

To address these challenges, we start from the intrinsic scene consistency of multi-modal remote sensing data, recognizing that different modalities can be regarded as modality-specific observations or representations of the same underlying scene. This insight leads us to reformulate multi-modal generation in a scene-centered paradigm, anchoring the generative process around the intrinsic scene content. As shown in Fig.~\ref{fig:teaser}, under this scene-centered joint modeling paradigm, a latent scene representation is first inferred from available observations, and target modalities are then generated conditioned on that representation. This paradigm also provides a unified foundation for both paired joint generation and cross-modal translation.

Building upon this paradigm, we propose MetaEarth-MM, a unified generative foundation model for multi-modal remote sensing imagery across five modalities: RGB, SAR, NIR, Panchromatic image (PAN), and OSM. To instantiate the proposed scene-centered joint modeling paradigm, MetaEarth-MM introduces a decoupled architecture for modeling the joint distribution of aligned multi-modal remote sensing images under the flow matching framework. Specifically, the model consists of two components: a scene inference module and a modality-aware routed generator. The scene inference module estimates a latent scene representation from available noisy modality latents. Conditioned on this inferred representation, the modality-aware routed generator predicts the velocity field of each target modality through modality-specific generation routes. Crucially, we further introduce a scene consistency regularization during training, encouraging different modal observations of the same scene to yield a latent representation centered on the underlying scene. In this way, MetaEarth-MM separates shared scene modeling from modality-specific generation, enabling diverse generation tasks within a unified model while reducing cross-modal interference and capacity competition caused by heterogeneous modality distributions.

To support unified multi-modal generation, we construct a large-scale multi-modal remote sensing dataset EarthMM. Existing datasets are often limited in modality diversity, dataset scale, or modality combinations, which restricts the development of unified multi-modal generative models. EarthMM contains 2.8 million images across five modalities, covering diverse geographic regions and a wide range of spatial resolutions from 0.5 to 10 m per pixel, thereby providing the data foundation for joint generative modeling across diverse modalities.

Comprehensive experiments demonstrate that MetaEarth-MM achieves superior generative performance in both paired joint generation and any-to-any translation across diverse modalities. It also exhibits strong zero-shot generalization, successfully synthesizing unseen modality combinations beyond those observed during training. Beyond image generation, MetaEarth-MM supports downstream tasks through generative data augmentation, image-level domain adaptation, and zero-shot representation transfer. These results suggest that MetaEarth-MM shows promise as both a powerful multi-modal generator and a versatile foundation model for multi-modal Earth observation.

The contributions of this paper are summarized as follows:

\begin{itemize}
\item
We propose MetaEarth-MM, a unified foundation model for multi-modal remote sensing image generation under a scene-centered joint modeling paradigm. Its decoupled architecture separates latent scene inference from modality-specific generation, enabling paired joint generation and any-to-any cross-modal translation across five modalities within a unified model.

\item
We construct EarthMM, a globally distributed multi-modal remote sensing dataset comprising 2.8 million images and 2.2 million aligned pairs. With larger-scale paired observations, broader modality diversity, and a wider range of spatial resolution than existing datasets, EarthMM serves as a comprehensive data foundation for developing multi-modal generative models.

\item
We demonstrate the downstream utility of MetaEarth-MM at both data and representation levels, showing its role as a multi-modal generative data engine and its ability to support zero-shot transfer through learned scene representations.
\end{itemize}

\section{Related Work}
\subsection{Unified Multi-modal Image Generation in General Vision}
Synthesizing images across diverse visual modalities remains a fundamental challenge in computer vision. Early efforts in cross-modal generation primarily focused on a collection of isolated Image-to-Image (I2I) translation tasks. Early GAN-based methods, such as Pix2Pix\cite{pix2pix, pix2pixhd} and CycleGAN\cite{cyclegan}, independently addressed different translation tasks, requiring a dedicated model for each pair of specific modality. Subsequent variants\cite{stargan, CUT, stegogan, qi2024layered} focused on enhancing synthesis fidelity and mapping robustness across more complex scenarios. 

With the advent of diffusion models, this fragmented I2I paradigm evolved toward unified multi-modal architectures. Seminal works like Palette \cite{palette} demonstrated that a single diffusion model can master diverse conditional generation problems. Following this trend, recent generative foundation models have further broadened the scope to support generation and translation tasks across an expanded set of visual modalities. These models\cite{instructpix2pix, magicbrush, omnigen2, pixwizard} can simultaneously handle bidirectional mappings between RGB images and various spatial conditions, encompassing tasks such as RGB-to-canny, RGB-to-pose and their inverse generation processes.

In parallel, significant progress has been made in adapting pre-trained foundation models for conditional cross-modal image generation. ControlNet\cite{controlnet} and its concurrent works\cite{adapter, composer} introduce parameter-efficient mechanisms to inject diverse visual modality guidance, such as edge maps and semantic layouts, into pre-trained generative foundation models. More recently, OminiControl\cite{ominicontrol} also extended these parameter-efficient condition injection mechanisms to modern DiT models\cite{flux, dit, sd3}. Building upon these frameworks, UniControl\cite{unicontrol} and Uni-ControlNet\cite{unicontrolnet} have demonstrated remarkable capabilities in handling multiple visual representations simultaneously, enabling more flexible and composable cross-modal image generation.

While successful in general vision, these methods typically treat image generation as an RGB-centric task where other modalities serve merely as auxiliary conditions. This primary-auxiliary assumption is ill-suited for Earth observation. Remote sensing data consists of parallel, equally significant observations of the identical geographical area. Thus, instead of using one modality simply to condition another, there is a critical need to explicitly model the joint distribution across these diverse modalities.

\subsection{Multi-modal Image Generation in Remote Sensing}
Historically, early research on multi-modal remote sensing image generation was largely developed in a modality-specific manner, with most methods designed for individual modalities or narrowly defined generation settings. One important line of work focuses on modality-specific generation conditioned on auxiliary signals other than source images, such as semantic descriptors, category information, or modality-dependent physical cues. In this setting, models are often designed to reflect the imaging principles of a particular sensor, enabling physics-informed and interpretable generation \cite{ph-gan, 9461232, 10752552, liu2022physics, liu2023diverse,Chen2024Spectral}. While these methods are effective at modeling the data distribution within a single modality, their specialized architectures and sensor-dependent assumptions make them difficult to extend toward unified cross-modal generation.

Another major line of research studies cross-modal image translation between heterogeneous sensor observations. Representative tasks include SAR-to-optical translation and NIR/PAN colorization. Most early methods were developed under an image-to-image translation framework (e.g., Pix2Pix~\cite{pix2pix}), and improved synthesis fidelity and spatial consistency through enhanced feature representations \cite{HybridcGAN}, frequency-domain or multi-scale constraints \cite{WFLM-GAN}, and temporal modeling \cite{MTS2ONet}. Subsequent work further addressed unpaired or data-limited settings by introducing priors from pre-trained optical foundation models. More recently, diffusion-based methods have also been adapted to remote sensing translation, particularly by incorporating inductive biases tailored to SAR characteristics \cite{SFDiff, E3Diff, CM-Diffusion}.

A related yet distinct direction is map-satellite translation, which connects abstract semantic layouts with realistic remote sensing imagery. Benefiting from the controllability of ControlNet and the strong priors of pre-trained diffusion models, methods such as Map-Sat \cite{map-sat} and GeoSynth \cite{geosynth} generate high-resolution remote sensing imagery from map layouts. Extending this direction, EarthMapper \cite{earthmapper} further explores bidirectional map-satellite translation and investigates the interaction between structured map representations and satellite observations.

While these methods are effective within their targeted tasks, most existing approaches remain centered on isolated conditional generation or pairwise image-to-image translation. Such formulations are typically designed for a single source-target mapping and therefore do not explicitly model the broader relationships among multiple remote sensing modalities. Although recent studies have begun to explore unified multi-modal generative modeling \cite{cop-gen-beta}, they typically rely on direct joint modeling paradigms that implicitly entangle diverse modalities. In contrast, MetaEarth-MM reformulates this process through a novel scene-centered paradigm. By introducing a latent scene representation as an intermediate variable, our method organizes the generative process around the intrinsic scene content.

\subsection{Remote Sensing Generative Foundation Model}
The rapid evolution of generative AI has facilitated the introduction of generative foundation models to remote sensing\cite{rsfdm, changen2, noise2change,diffusionsat,rsdiff,liu2023decoupling}. Early efforts in this area mainly focused on adapting latent diffusion models\cite{sd} for remote sensing text-to-image generation. Representative works such as DiffusionSat\cite{diffusionsat} and RSDiff\cite{rsdiff} demonstrated the feasibility of adapting established natural image generative architectures\cite{imagen} to remote sensing imagery.

Building on these advances, subsequent studies further expanded both the scale and scope of remote sensing generation. MetaEarth \cite{metaearth, metaearth3d} introduced a resolution-guided self-cascading framework that enables unbounded multi-resolution remote sensing image generation at a global scale. In parallel, Text2Earth\cite{text2earth} established Git-10M, a global-scale image-text pair dataset, to train billion-parameter foundation models, which demonstrates remarkable flexibility across diverse generative tasks, including zero-shot generation, image editing, and cross-modal image translation. 

To provide these foundation models with finer spatial and structural control, later research increasingly turned to more precise conditional generation. A prominent line of work focuses on adapting ControlNet\cite{controlnet} or T2I-Adapters\cite{adapter} to remote sensing, employing dense structural inputs, such as semantic maps, layouts, or edge images, to guide the generation process. For instance, Crs-Diff\cite{crsdiff} supports multiple control signals for guided image generation, while HSIGene\cite{hsigene} further extends this multi-conditional framework to hyperspectral imagery. Beyond external control branches, other studies refine internal model representations to improve controllability. A representative example is MGDiff \cite{mgdiff}, which enables precise object-count control by imposing constraints at three granularities.

Despite the substantial progress of existing image-conditioned remote sensing generative foundation models, their primary focus has largely remained on visible-image generation and spatially controlled synthesis. On the one hand, many current methods build upon backbones pretrained on RGB imagery, and their generation space is correspondingly centered on visible optical outputs. On the other hand, their conditioning inputs are typically general spatial layouts, which are less directly suited to modeling the translation relationships among diverse Earth observation modalities. In this paper, we propose MetaEarth-MM, a unified remote sensing generative foundation model that extends generative modeling beyond visible imagery and supports multi-modal generation across five modalities. Within a single framework, it enables joint generation and zero-shot any-to-any translation, broadening the scope of remote sensing generative foundation models.

\section{The proposed MetaEarth-MM}

\begin{figure*}[t]
    \centering
    \includegraphics[width=\textwidth]{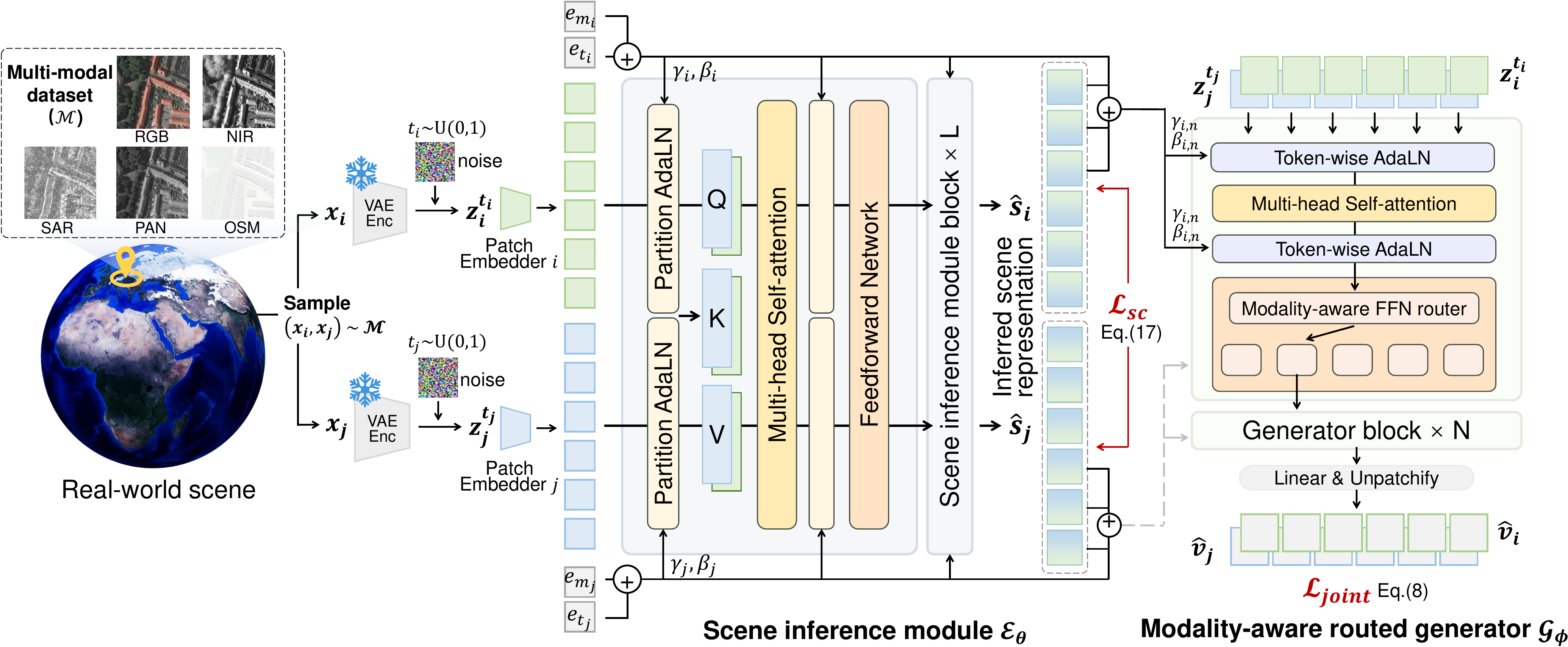}
    \caption{\textbf{Overall architecture of MetaEarth-MM.} The model adopts a decoupled architecture consisting of two components: a scene inference module that infers the latent scene representation from paired noisy observations, and a modality-aware routed generator that predicts modality-specific velocity fields conditioned on the inferred scene representation. During training, a scene consistency regularization is further introduced, encouraging different modal observations of the same scene to yield a latent representation centered on the underlying scene.}
    \label{fig:method}
\end{figure*}

\subsection{From conditional to joint flow matching}
Most existing methods achieve paired multi-modal image generation through isolated cross-modal translation. Given a set of remote sensing modalities $\mathcal{M}$, e.g., $\mathcal{M}=\{\text{RGB}, \text{NIR}, \text{SAR}, \text{PAN}, \text{OSM}\}$, let $\mathbf{x}_i$ and $\mathbf{x}_j$ denote paired observations from two different modalities $m_i, m_j \in \mathcal{M}$, they typically learn a conditional distribution $p(\mathbf{x}_i | \mathbf{x}_j)$. However, this paradigm confines paired generation to scenarios where a source modality is observed, and scales poorly with modality diversity. To overcome these limitations, we shift our focus to joint distribution modeling. By learning the joint distribution $p(\mathbf{x}_i,\mathbf{x}_j)$, the model subsumes pairwise conditional generation while naturally enabling source-free joint sampling.

To accomplish this modeling process, we adopt the continuous flow matching framework~\cite{fm}. In the following, we detail the mathematical transition from conditional to joint flow matching. For a modality $m_i$, let $\mathbf{z}_i^0$ denote the latent embedding of $\mathbf{x}_i$ encoded by a pre-trained variational autoencoder (VAE)~\cite{sd}, and let $\mathbf{z}_i^1 \sim \mathcal{N}(\mathbf{0}, \mathbf{I})$ be a standard Gaussian sample. Rectified flow~\cite{rectified_flow} defines a linear interpolation path between data and noise:
\begin{equation}
    \mathbf{z}_i^t = (1-t)\mathbf{z}_i^0 + t\mathbf{z}_i^1
\end{equation}
where $t \in [0,1]$ is the continuous time variable. The corresponding target velocity field is given by
\begin{equation}
    \mathbf{v}_i = \frac{\mathrm{d}\mathbf{z}_i^t}{\mathrm{d}t} = \mathbf{z}_i^1 - \mathbf{z}_i^0
\end{equation}

Under the conditional distribution modeling paradigm, the model only predicts the velocity of the target modality while treating the other modality as a clean condition. Specifically, for generating modality $m_i$ conditioned on $m_j$, the model learns
\begin{equation}
    \hat{\mathbf{v}}_i = f_\theta(\mathbf{z}_i^t, t, \mathbf{z}_j^0),
\end{equation}
and is trained with the standard flow matching objective:
\begin{equation}
    \mathcal{L}_{\text{cond}}
    = \mathbb{E}_{\mathbf{z}_i^0,\mathbf{z}_j^0,\mathbf{z}_i^1,t}
    \left[
    \left\|
    f_\theta(\mathbf{z}_i^t, t, \mathbf{z}_j^0) - \mathbf{v}_i
    \right\|_2^2
    \right].
\end{equation}

To support a more comprehensive joint generative framework, recent studies~\cite{unidiffuser, omniflow} have shown that multiple modalities can be jointly modeled by independently perturbing each modality with its own noise level. Following this paradigm, we independently corrupt the latent representations of both modalities and obtain the joint noisy state:

\begin{equation}
    \mathbf{z}_i^{t_i} = (1-t_i)\mathbf{z}_i^0 + t_i\mathbf{z}_i^1,
    \qquad
    \mathbf{z}_j^{t_j} = (1-t_j)\mathbf{z}_j^0 + t_j\mathbf{z}_j^1
\end{equation}
where $t_i, t_j \in [0,1]$ are independently sampled timesteps. Their corresponding target velocities are
\begin{equation}
    \mathbf{v}_i = \mathbf{z}_i^1 - \mathbf{z}_i^0,
    \qquad
    \mathbf{v}_j = \mathbf{z}_j^1 - \mathbf{z}_j^0
\end{equation}
The model then jointly predicts the dual velocity fields from the noisy modality pair:
\begin{equation}
    (\hat{\mathbf{v}}_i, \hat{\mathbf{v}}_j)
    =
    F_\theta(\mathbf{z}_i^{t_i}, \mathbf{z}_j^{t_j}, t_i, t_j)
\end{equation}
with the training objective
\begin{equation}
    \mathcal{L}_{\text{joint}}
    =
    \mathbb{E}
    \left[
    \left\|
    \hat{\mathbf{v}}_i - \mathbf{v}_i
    \right\|_2^2
    +
    \left\|
    \hat{\mathbf{v}}_j - \mathbf{v}_j
    \right\|_2^2
    \right].
\end{equation}

By sampling $t_i$ and $t_j$ independently during training, a single model can jointly learn marginal, conditional, and joint generation within a unified framework. This joint modeling paradigm provides the mathematical basis for unified multi-modal generation. we next present our scene-centered joint modeling paradigm for unified multi-modal remote sensing image generation.

\subsection{Overview of the MetaEarth-MM scene-centered joint modeling paradigm}
Building upon the joint flow matching framework, we propose a scene-centered joint modeling paradigm for unified multi-modal remote sensing generation. Driven by the intrinsic scene consistency of multi-modal data, we organize the generative process around the underlying scene content. Based on this paradigm, we develop a unified generative foundation model MetaEarth-MM with a decoupled architecture. Fig.~\ref{fig:method} illustrates the overall architecture of the proposed model. Specifically, the model first infers a latent scene representation from current noisy observations, and then uses this representation to guide the generation of each modality with its own observation and representation characteristics.

Given a paired sample $(\mathbf{x}_i, \mathbf{x}_j)$ from modalities $m_i$ and $m_j$, we assume that both observations correspond to the same underlying scene state, denoted by $\mathbf{s}$. Here, $\mathbf{s}$ is a conceptual variable representing the scene underlying the paired observations. Under this scene-centered paradigm, the joint modeling of the modality pair is explicitly decoupled by conditioning on this shared scene state:
\begin{equation}
    p(\mathbf{x}_i, \mathbf{x}_j | \mathbf{s}) = p(\mathbf{x}_i | \mathbf{s}) \, p(\mathbf{x}_j | \mathbf{s})
\end{equation}

This conditional independence directly motivates our decoupled architectural design: given the inferred scene state, the generation of each modality can be carried out independently. Correspondingly, within the flow matching framework, the joint velocity field $(\hat{\mathbf{v}}_i, \hat{\mathbf{v}}_j)$ naturally decouples into independent modality-specific velocity predictions conditioned on this shared scene state.

However, since the state $\mathbf{s}$ is unobserved in practice, it is not directly accessible to the model. To address this, we design a scene inference module $\mathcal{E}_{\theta}$ to implicitly estimate latent scene representations from noisy observations. Given a pair of noisy modality latents $(\mathbf{z}_i^{t_i}, \mathbf{z}_j^{t_j})$, the scene inference module processes them jointly and outputs paired scene embeddings, denoted as $\hat{\mathbf{s}}_i$ and $\hat{\mathbf{s}}_j$: 

\begin{equation}
    [\hat{\mathbf{s}}_i, \hat{\mathbf{s}}_j] =
    \mathcal{E}_{\theta}(\mathbf{z}_i^{t_i}, \mathbf{z}_j^{t_j}, t_i, t_j, m_i, m_j)
\end{equation}
where $m_i$ and $m_j$ denote modality identifiers. Since both embeddings are inferred from observations of the same scene, we further introduce a scene-consistency regularization during training to encourage $\hat{\mathbf{s}}_i$ and $\hat{\mathbf{s}}_j$ to remain close, so that they can better approximate the same underlying scene state $\mathbf{s}$. Details are provided in Sec.~IV-C.

Conditioned on the inferred latent scene representation, we further design a unified modality-aware routed generator $\mathcal{G}_{\phi}$ to predict the target velocity field for each modality independently:
\begin{equation}
    \hat{\mathbf{v}}_i = \mathcal{G}_{\phi}(\mathbf{z}_i^{t_i}, \hat{\mathbf{s}}_i, t_i, m_i), \qquad
    \hat{\mathbf{v}}_j = \mathcal{G}_{\phi}(\mathbf{z}_j^{t_j}, \hat{\mathbf{s}}_j, t_j, m_j)
\end{equation}
where $\hat{\mathbf{v}}_i$ and $\hat{\mathbf{v}}_j$ denote the predicted velocity fields for modalities $m_i$ and $m_j$, respectively. Structurally, this decoupled model design organizes cross-modal interaction at the scene inference stage, while leaving modality-specific velocity prediction to separate generation processes, thereby helping alleviate representation entanglement across heterogeneous modalities.

\subsection{Scene inference module}
The scene inference module $\mathcal{E}_{\theta}$ is designed to estimate a latent scene representation from the paired noisy observations. To this end, we build $\mathcal{E}_{\theta}$ upon a diffusion transformer (DiT) architecture and adapt it to the setting where paired modalities are jointly processed with independently sampled noise levels.

For each modality $k \in \{i,j\}$, the noisy latent $\mathbf{z}_k^{t_k} \in \mathbb{R}^{H \times W \times C}$ is first converted into a sequence of $N$ tokens with dimension $D$ through a modality-specific patch embedding layer. Since the paired modalities are spatially aligned, we inject a shared spatial positional embedding $\mathbf{E}_{pos} \in \mathbb{R}^{N\times D}$ into different modality token sequences to preserve location-wise correspondence. Meanwhile, a learnable modality embedding $\mathbf{e}_{m_k} \in \mathbb{R}^{D}$ is added to retain modality identity. The resulting input tokens are written as:
\begin{equation}
\mathbf{H}_k^{(0)} = \mathrm{Patch}_k(\mathbf{z}_k^{t_k}) + \mathbf{E}_{pos} + \mathbf{e}_{m_k}
\label{eq:patchify}
\end{equation}
and these tokens from both modalities are concatenated along sequence dimension to form the joint input sequence $\mathbf{H}^{(0)}=[\mathbf{H}_i^{(0)}; \mathbf{H}_j^{(0)}]\in \mathbb{R}^{2N\times D}$.

The $\mathbf{H}^{(0)}$ is subsequently processed by $L$ standard DiT blocks. Within the $l$-th block, different modality partitions share the same set of attention projection layers (i.e., $\mathbf{W}_q^l, \mathbf{W}_k^l, \mathbf{W}_v^l \in \mathbb{R}^{D \times D}$). The self-attention is then performed over the full sequence, allowing each token to aggregate both intra-modal context and cross-modal correspondences. Meanwhile, since the tokens of each modality are perturbed under independently sampled noise levels and exhibit distinct modality characteristics, we introduce a partition adaptive layer normalization to enable the model to adapt to differences in both noise scale and modality-specific features. Specifically, for each modality $k \in \{i,j\}$, the modulation parameters $\boldsymbol{\gamma}_k^{(l)}$ and $\boldsymbol{\beta}_k^{(l)}$ are generated from a condition embedding $\mathbf{c}_k=\mathbf{e}_{t_k} + \mathbf{e}_{m_k}$, where $\mathbf{e}_{t_k}\in\mathbb{R}^{D}$ denotes the embedding of timestep $t_k$:

\begin{equation}
[\boldsymbol{\gamma}_k^{(l)}, \boldsymbol{\beta}_k^{(l)}] = \mathrm{MLP}(\mathbf{c}_k)
\end{equation}
where MLP denotes a multi-layer perceptron. These parameters are applied only to the tokens of the corresponding modality partition:
\begin{equation}
\bar{\mathbf{H}}_k^{(l-1)}=\boldsymbol{\gamma}_k^{(l)} \odot \mathrm{LayerNorm}(\mathbf{H}_k^{(l-1)}) + \boldsymbol{\beta}_k^{(l)}
\label{eq:adaln}
\end{equation}
where $\odot$ represents element-wise multiplication. In this way, pairwise dependencies are jointly modeled in a shared attention space, while the network remains flexible to accommodate heterogeneous feature patterns across modalities.

After $L$ blocks, the final sequence is denoted as $\mathbf{H}^{(L)}=[\hat{\mathbf{s}}_i, \hat{\mathbf{s}}_j]$, where $\hat{\mathbf{s}}_i, \hat{\mathbf{s}}_j \in \mathbb{R}^{N\times D}$ represent the latent scene embeddings for each modality. Since both embeddings serve as estimates of the same underlying scene, we further impose a scene-consistency regularization. Specifically, the similarity between $\hat{\mathbf{s}}_i$ and $\hat{\mathbf{s}}_j$ is defined by averaging the cosine similarity over spatially corresponding tokens:
\begin{equation}
\mathrm{sim}(\hat{\mathbf{s}}_i, \hat{\mathbf{s}}_j) = \frac{1}{N} \sum_{n=1}^{N} \frac{\hat{\mathbf{s}}_{i,n}^{\top} \hat{\mathbf{s}}_{j,n}}{\|\hat{\mathbf{s}}_{i,n}\|_2 \, \|\hat{\mathbf{s}}_{j,n}\|_2}
\end{equation}
where $\hat{\mathbf{s}}_{i,n}, \hat{\mathbf{s}}_{j,n} \in \mathbb{R}^{D}$ denote the $n$-th tokens in the embeddings $\hat{\mathbf{s}}_i$ and $\hat{\mathbf{s}}_j$. Based on this similarity, we define the scene-consistency loss using a symmetric InfoNCE objective over a mini-batch of size $B$. The directional loss from modality $i$ to $j$ is defined as follows:
\begin{equation}
\mathcal{L}_{\mathrm{sc}}^{i \rightarrow j}
=-\frac{1}{B}\sum_{b=1}^{B} \log \frac{\exp(\mathrm{sim}(\hat{\mathbf{s}}_i^{(b)}, \hat{\mathbf{s}}_j^{(b)})/\tau)}{\sum_{m=1}^{B} \exp(\mathrm{sim}(\hat{\mathbf{s}}_i^{(b)}, \hat{\mathbf{s}}_j^{(m)})/\tau)}
\end{equation}
where $\tau$ is a learnable temperature parameter, $\hat{\mathbf{s}}_i^{(b)}$ and $\hat{\mathbf{s}}_j^{(b)}$ denote the paired scene embeddings of the $b$-th sample in the mini-batch. The full scene-consistency loss is then given by the average of the two directional InfoNCE terms:

\begin{equation}
\mathcal{L}_{\mathrm{sc}}=\frac{1}{2}\left(\mathcal{L}_{\mathrm{sc}}^{i \rightarrow j}+\mathcal{L}_{\mathrm{sc}}^{j \rightarrow i}
\right)
\end{equation}

\subsection{Modality-aware routed generator}
The modality-aware routed generator $\mathcal{G}_{\phi}$ is designed to predict the latent velocity field of each modality conditioned on the inferred scene representation. The generator $\mathcal{G}_{\phi}$ is also built on a DiT-style architecture. For a modality $m_k$, where $k \in \{i,j\}$, the noisy latent $\mathbf{z}_k^{t_k}$ is first patchified into a sequence of latent tokens $\mathbf{U}_k^{(0)} \in \mathbb{R}^{N\times D}$ with its own modality-specific patch embedder. Positional and modality embeddings are added in the same way as Eq. ~\ref{eq:patchify}. In the $l$-th block, the latent scene representation is injected via token-wise AdaLN, providing spatially aligned guidance for the denoising process. Specifically, let $\hat{\mathbf{s}}_k = [\hat{\mathbf{s}}_{k,1}, \ldots, \hat{\mathbf{s}}_{k,N}]$ denote the inferred scene token sequence. For the $n$-th token $\mathbf{u}_{k,n}^{(l-1)} \in \mathbf{U}_k^{(l-1)}$ of modality $m_k$, the modulation parameters are derived from $\mathbf{c}_{k,n} = \hat{\mathbf{s}}_{k,n} + \mathbf{e}_{t_k} + \mathbf{e}_{m_k}$ and applied following the AdaLN formulation in Eq.~\ref{eq:adaln}. 

To reduce cross-modal interference and network capacity competition in unified multimodal generation, we further introduce deterministic modality routing in the FFN layers, while keeping self-attention shared across modalities. This design follows the different functional roles of the two components: self-attention mainly captures spatial and contextual dependencies that can be shared across modalities, whereas the FFN is more directly responsible for modeling modality-specific features. Specifically, the FFN in the $l$-th block is implemented with a set of routed branches $\{\mathcal{F}_{r}^{(l)}\}_{r=1}^{R}$. For the modality $m_k$, a fixed routing indicator $\boldsymbol{\pi}(m_k) \in \{0,1\}^{R}$ is determined by its modality identity, with $\sum_{r=1}^{R}\pi_r(m_k)=1$. The FFN operation can be written as
\begin{equation}
\mathrm{FFN}^{(l)}(\mathbf{U}_k^{(l-1)}, m_k)=\sum_{r=1}^{R}
\pi_r(m_k)\,\mathcal{F}_{r}^{(l)}(\mathbf{U}_k^{(l-1)})
\end{equation}
Since $\boldsymbol{\pi}(m_k)$ is one-hot, all tokens from the same modality are routed to the same branch. After passing through all transformer blocks, the output tokens are projected and unpatchified to produce the predicted velocity field.

With the scene inference module and the modality-aware routed generator designed above, MetaEarth-MM is trained end-to-end with a joint objective that combines the flow matching loss and the scene-consistency regularization:
\begin{equation}
\mathcal{L}=\mathcal{L}_{\text{joint}}+\lambda\mathcal{L}_{\mathrm{sc}}
\label{eq:overall_loss}
\end{equation}
where $\lambda$ is a balancing coefficient.

\section{Experiments}

\subsection{The EarthMM dataset}
To support unified multimodal remote sensing generation, we construct EarthMM, a large-scale dataset spanning five modalities: RGB, SAR, NIR, PAN, and OSM. This dataset contains 2.8 million globally distributed images and spans a wide range of resolutions, from 0.5 to 10 m per pixel. Table~\ref{tab:dataset_comparison} compares EarthMM with representative remote sensing datasets, showing that EarthMM provides broader modality coverage, larger scale, and a wider resolution range.

\begin{table}[t]
\centering
\setlength{\tabcolsep}{3.2pt}
\renewcommand{\arraystretch}{1.05}
\caption{Comparison of EarthMM with representative multimodal remote sensing datasets. Image and image-pair counts are reported after normalizing the original image sizes to 256$\times$256 patches.}
\label{tab:dataset_comparison}
\resizebox{\columnwidth}{!}{
\begin{tabular}{ccccccccc}
\toprule
\multirow{2}{*}{Dataset} & \multirow{2}{*}{\#Images} & \multirow{2}{*}{\#Pairs} & \multicolumn{5}{c}{Modalities} & \multirow{2}{*}{\shortstack{Spatial \\resolution (m/pix)}} \\
\cmidrule(lr){4-8}
& & & RGB & SAR & NIR & OSM & PAN & \\
\midrule
SEN1-2~\cite{sen1-2}& 564K& 282K& \checkmark & \checkmark &  &  &   & 10 \\
SAR2Opt~\cite{sar2opt}& 22K& 11K& \checkmark & \checkmark &  &  &   & 1 \\
MMM-RS~\cite{mmm-rs}& 594K& 297K& \checkmark & \checkmark & \checkmark  &  &   & 5-10 \\
MSBC~\cite{msbc}& 11.4K& 3.8K& \checkmark & \checkmark & \checkmark  &  &   & 2 \\
S2MS-HR~\cite{s2ms-hr}& 9.1K& 4.5K&  & \checkmark & \checkmark  &  &   & 0.3 \\
GID~\cite{gid}& 237K& 118K& \checkmark&  & \checkmark & & & 4 \\
CNSatMap~\cite{earthmapper}& 604K& 302K& \checkmark & & &\checkmark & &0.6 \\
NewYork Map~\cite{pix2pix}& 36K& 18K& \checkmark & & &\checkmark & & - \\
\midrule
EarthMM (Ours) & 2.8M & 2.2M & \checkmark & \checkmark & \checkmark & \checkmark & \checkmark & 0.5-10 \\
\bottomrule
\end{tabular}
}
\end{table}

\subsubsection{Data collection and preprocessing} EarthMM is built by combining 15 public high-quality pair-wise aligned remote sensing datasets with a newly collected set of RGB-OSM pairs. Specifically, the integrated multi-modal datasets include OpenEarthMap-SAR~\cite{openearthmap}, DFC23~\cite{dfc23}, DDHRNet~\cite{ddhrnet}, fMoW~\cite{fmow}, MultiResSAR~\cite{multiressar}, OSDataset~\cite{osdataset}, QXS-SAROPT~\cite{qsxsaropt}, SEN12MS~\cite{sen12ms}, SpaceNet-3~\cite{spacenet35}, SpaceNet-5~\cite{spacenet35}, SpaceNet6~\cite{spacenet6}, SSL4EO-S12~\cite{ssl4eo}, WHU-OPT-SAR~\cite{whu}, and 3MOS~\cite{3mos}. We further apply strict and consistent data cleaning process across all sources, removing samples affected by severe noise, corrupted observations, or large-area cloud occlusion. Beyond the integrated public datasets, we further collect a large-scale set of geographically aligned RGB-OSM pairs covering 42 representative cities worldwide, including London, Beijing, New York, Tokyo. The RGB images are sourced from Google Earth, while the OSM data are obtained from OpenStreetMap and rendered into RGB-style map images. Each image-map pair is aligned by geographic coordinates to ensure spatial correspondence. To improve diversity and reduce redundancy, we remove highly repetitive regions, such as homogeneous bare-land areas. All data are finally cropped or resampled into uniform 256$\times$256 patches for training and analysis.

\begin{figure}[t]
    \centering
    \includegraphics[width=\linewidth]{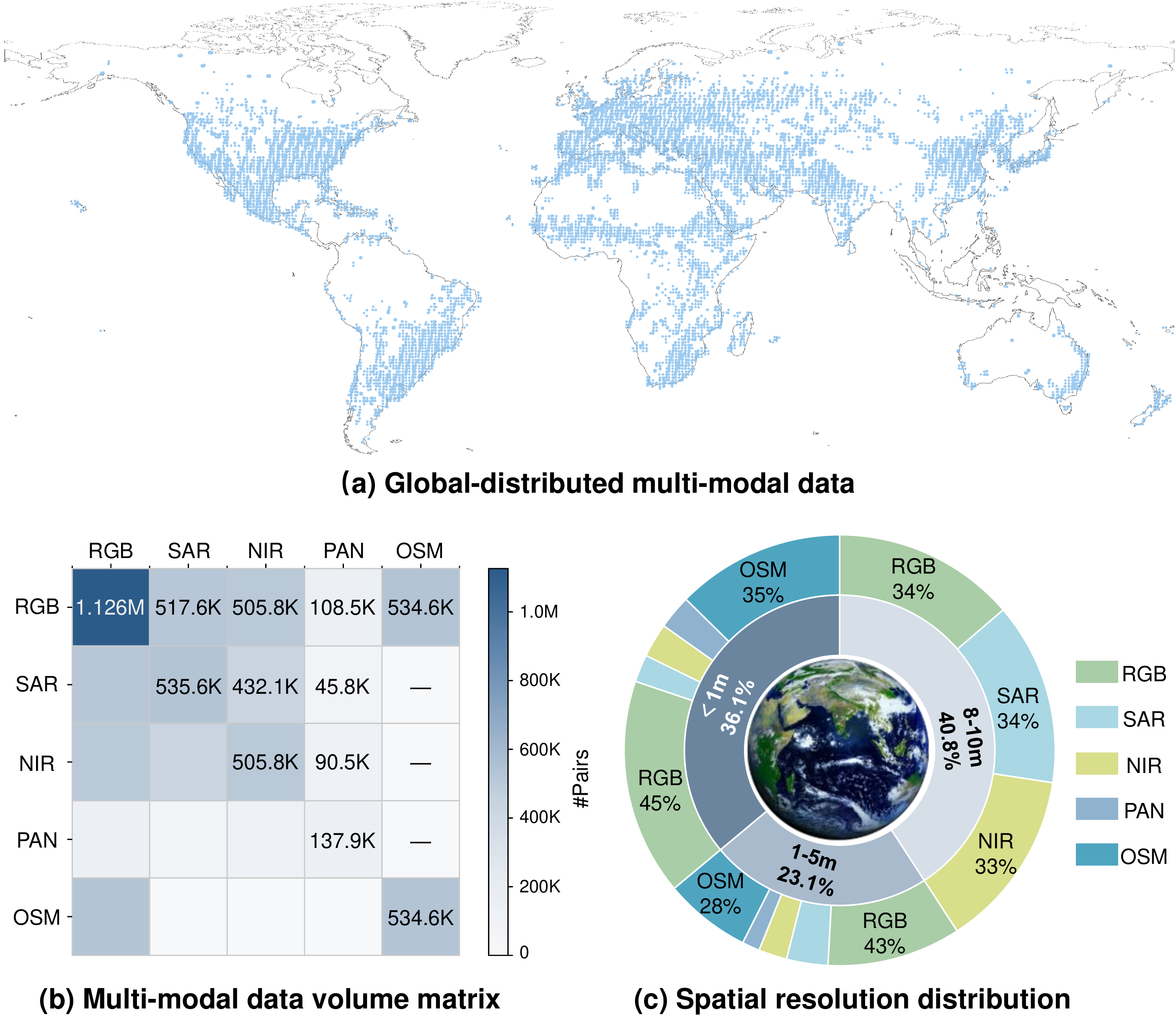}
    \caption{\textbf{Overview of the EarthMM dataset.} (a) Global geographic distribution of collected multi-modal samples. (b) Multi-modal data volume matrix, where diagonal entries denote the number of samples for each modality and off-diagonal entries denote the number of aligned modality pairs. (c) Spatial resolution distribution and modality composition across different resolution ranges.}
    \label{fig:dataset}
\end{figure}

\subsubsection{Dataset analysis} Fig.~\ref{fig:dataset} summarizes the data distribution of EarthMM from three aspects, including geographic coverage, cross-modal pair statistics, and spatial resolution composition. As shown in Fig.~\ref{fig:dataset}(a), the dataset covers a wide range of land areas across major continents, suggesting its diversity in geographic environments, land-cover patterns, and scene layouts. This broad distribution is beneficial for developing multimodal generative models with stronger scene diversity and better geographic generalization. Fig.~\ref{fig:dataset}(b) presents the sample counts of each modality and the numbers of aligned cross-modal pairs in EarthMM. The diagonal entries denote the sample counts of individual modalities, while the off-diagonal entries denote the numbers of aligned cross-modal pairs. In total, the dataset contains about 2.2M modality pairs, which provide rich cross-modal supervision for learning correspondences across diverse modality combinations and for capturing the underlying scene content. Fig.~\ref{fig:dataset}(c) further reveals how different modalities are distributed across spatial resolutions in EarthMM. The inner ring shows that the dataset covers three major resolution ranges, including $\leq$0.5 m/pix, 1-5 m/pix, and 8-10 m/pix, while the outer ring further presents the modality composition within each range. The RGB data span all three major resolution ranges and thus connect multiple spatial granularities, while OSM and PAN are mainly distributed in higher-resolution ranges with richer fine-scale structural details, and SAR and NIR are more prominent in the medium resolution range. This resolution composition enables EarthMM to cover both fine spatial structures and broader cross-sensor observations, which helps improve the generalization of the model across different observation scales.

\subsection{Experimental setup}
\subsubsection{Model design}
MetaEarth-MM is instantiated with a 20-block scene inference module and a 4-block modality-aware routed generator. Both modules use a hidden dimension of 1024. The latent inputs are patchified with a patch size of 2. In the routed generator, the number of FFN branches is set to 5, matching the number of modalities, and each modality is deterministically assigned to its corresponding branch. All modalities share the same pre-trained VAE for latent encoding, which is kept frozen during training. In total, the MetaEarth-MM contains approximately 600M parameters.

\subsubsection{Training details}
Our model is implemented in PyTorch and trained from scratch without any pre-trained initialization. We use the AdamW optimizer with an initial learning rate of $1\times10^{-5}$. Training is conducted on NVIDIA A800 GPUs with a total batch size of 256 for 900K iterations. To alleviate the imbalance among different modality pairs in the training set, we uniformly sample modality pairs during training. To enable classifier-free guidance during inference, we randomly drop one modality input with a probability of 0.1 during training, where the dropped modality is replaced with pure noise. In addition, we randomly convert RGB images into grayscale images to construct pseudo-PAN samples as an auxiliary augmentation for the PAN modality. The model is optimized using the overall objective in Eq.~\ref{eq:overall_loss}, with loss weight set to $\lambda=1$.

\subsubsection{Evaluation settings}
For evaluation, we construct test sets by sampling paired samples from EarthMM. The benchmark covers four modality pairs used in the main comparison, including SAR-RGB, NIR-RGB, PAN-RGB, OSM-RGB with 6826, 6670, 6670, and 10000 paired samples, respectively. The same test sets are used for all compared methods to ensure a fair evaluation.

During inference, MetaEarth-MM uses the Euler-Maruyama sampler with 250 sampling steps for all modality pairs. Classifier-free guidance is applied with a guidance scale of 1.2. For cross-modal translation, the source modality is fixed as the condition, and the target modality is generated from noise. For unconditional paired joint generation, both modalities are initialized from noise and generated jointly in a single sampling process.

\subsubsection{Evaluation metrics}
We evaluate the generated results comprehensively from four aspects: distributional realism, perceptual fidelity, structural similarity, and cross-modal consistency. Specifically, we adopt FID~\cite{fid} for distributional realism, LPIPS~\cite{lpips} for perceptual fidelity, and SSIM~\cite{ssim} for structural similarity. However, for SAR-related generative tasks, the inherent speckle noise makes local intensity statistics less stable, often compromising SSIM. To address this, we additionally report PSNR and FSIM~\cite{fsim} for SAR images. FSIM is based on phase congruency and gradient magnitude, and thus better captures structurally salient features such as edges and scattering patterns.

To measure cross-modal consistency, we introduce the Cross-modal Alignment Score (CAS). For each modality pair, we train a CLIP-style dual-image encoder on the corresponding paired data using a contrastive learning objective~\cite{clip}. Given $N$ generated samples, let $\hat{\mathbf{x}}_i^{(n)}$ denote the generated image from modality $m_i$, and $\mathbf{x}_j^{(n)}$ denote its corresponding image from the other modality $m_j$. CAS is computed as the average cosine similarity in the learned embedding space:
\begin{equation}
\mathrm{CAS}_{i,j}
=
100 \times \frac{1}{N}
\sum_{n=1}^{N}
\frac{
f_i(\hat{\mathbf{x}}_i^{(n)})^\top f_j(\mathbf{x}_j^{(n)})
}{
\left\| f_i(\hat{\mathbf{x}}_i^{(n)}) \right\|_2
\left\| f_j(\mathbf{x}_j^{(n)}) \right\|_2
},
\end{equation}
where \(f_i\) and \(f_j\) are the modality-specific image encoders. A higher CAS indicates better cross-modal alignment between the generated image.

\subsection{Qualitative analysis}

\begin{figure*}[t]
    \centering
    \includegraphics[width=\textwidth]{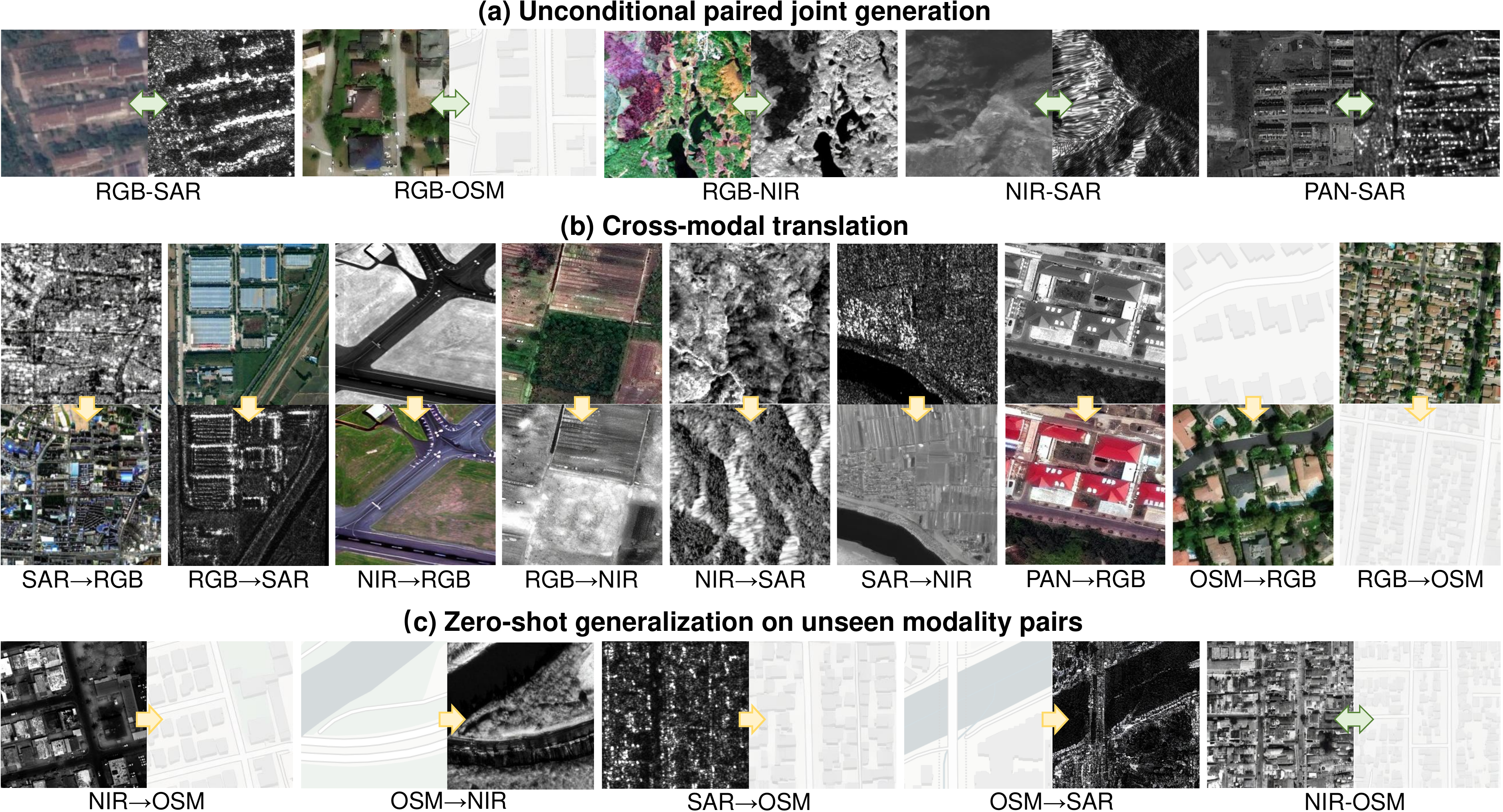}
    \caption{\textbf{Qualitative results of MetaEarth-MM under different generation settings.} (a) Examples of unconditional paired joint generation. (b) Examples of cross-modal translation. (c) Zero-shot generation on unseen generation tasks.}
    \label{fig:exp_multitask}
\end{figure*}

\subsubsection{Unified multimodal generation within a single framework}
Fig.~\ref{fig:exp_multitask}(a) and Fig.~\ref{fig:exp_multitask}(b) present qualitative results of MetaEarth-MM on two paired-modality generation tasks: unconditional paired joint generation and cross-modal translation, respectively. These results demonstrate that MetaEarth-MM serves as a unified generative foundation model capable of supporting both tasks without requiring task-specific model designs. In unconditional paired joint generation, the model generates two aligned modalities directly from noise. The generated RGB-SAR, RGB-OSM, RGB-NIR, NIR-SAR, and PAN-SAR pairs exhibit consistent scene structures across modalities. For cross-modal translation, the model supports multiple directions, as exemplified by SAR$\rightarrow$RGB, RGB$\rightarrow$SAR, NIR$\rightarrow$RGB, RGB$\rightarrow$NIR, NIR$\rightarrow$SAR, SAR$\rightarrow$NIR, PAN$\rightarrow$RGB, OSM$\rightarrow$RGB, and RGB$\rightarrow$OSM. The translated results preserve major spatial layouts, such as roads, buildings, vegetation, and field boundaries. Across both tasks, the generated modalities also show visually reasonable modality-dependent appearances: water areas are often represented as dark homogeneous regions in SAR, vegetation shows stronger responses in NIR and OSM maps assign correct colors to different land-cover categories.

\subsubsection{Zero-shot generation on unseen modality pairs}
Fig.~\ref{fig:exp_multitask}(c) further shows the zero-shot capability of MetaEarth-MM on unseen modality pairs. According to the modality-pair statistics of EarthMM in Fig.~\ref{fig:dataset}, OSM-NIR and OSM-SAR pairs are not included in the paired training data. Despite the absence of direct paired supervision, the model still generates visually plausible results with reasonable spatial and semantic correspondence. For example, the generated OSM maps exhibit precise scene layouts and color mappings, while the NIR and SAR images generated from OSM translate the category semantics of the map input into modality-specific appearances. For NIR-OSM joint generation, the paired outputs also exhibit coherent spatial layouts across modalities. These results suggest that the scene-centered joint modeling paradigm allows the model to reuse modality-scene relationships learned from available pairs and compose them for unseen modality combinations.

\subsection{Comparision with previous methods}
We compare MetaEarth-MM with representative previous methods under two tasks: cross-modal translation and unconditional paired joint generation. For cross-modal translation, we include Pix2Pix~\cite{pix2pix} as a GAN-based image-to-image translation method, Palette~\cite{palette} as a pixel-space diffusion baseline, and BBDM~\cite{bbdm}/BiBBDM~\cite{bibbdm} as latent-space Brownian bridge diffusion models. We also compare with Text2Earth~\cite{text2earth}, a remote sensing text-to-image diffusion model that supports modality translation through a ControlNet-based architecture~\cite{controlnet}. During implementation, we initialize these methods with their officially released pre-trained weights whenever available. Each method is trained independently for each specific task, while MetaEarth-MM is evaluated across all tasks without any task-specific fine-tuning. 

\subsubsection{Comparison on cross-modal translation}
\begin{figure*}[t]
    \centering
    \includegraphics[width=\textwidth]{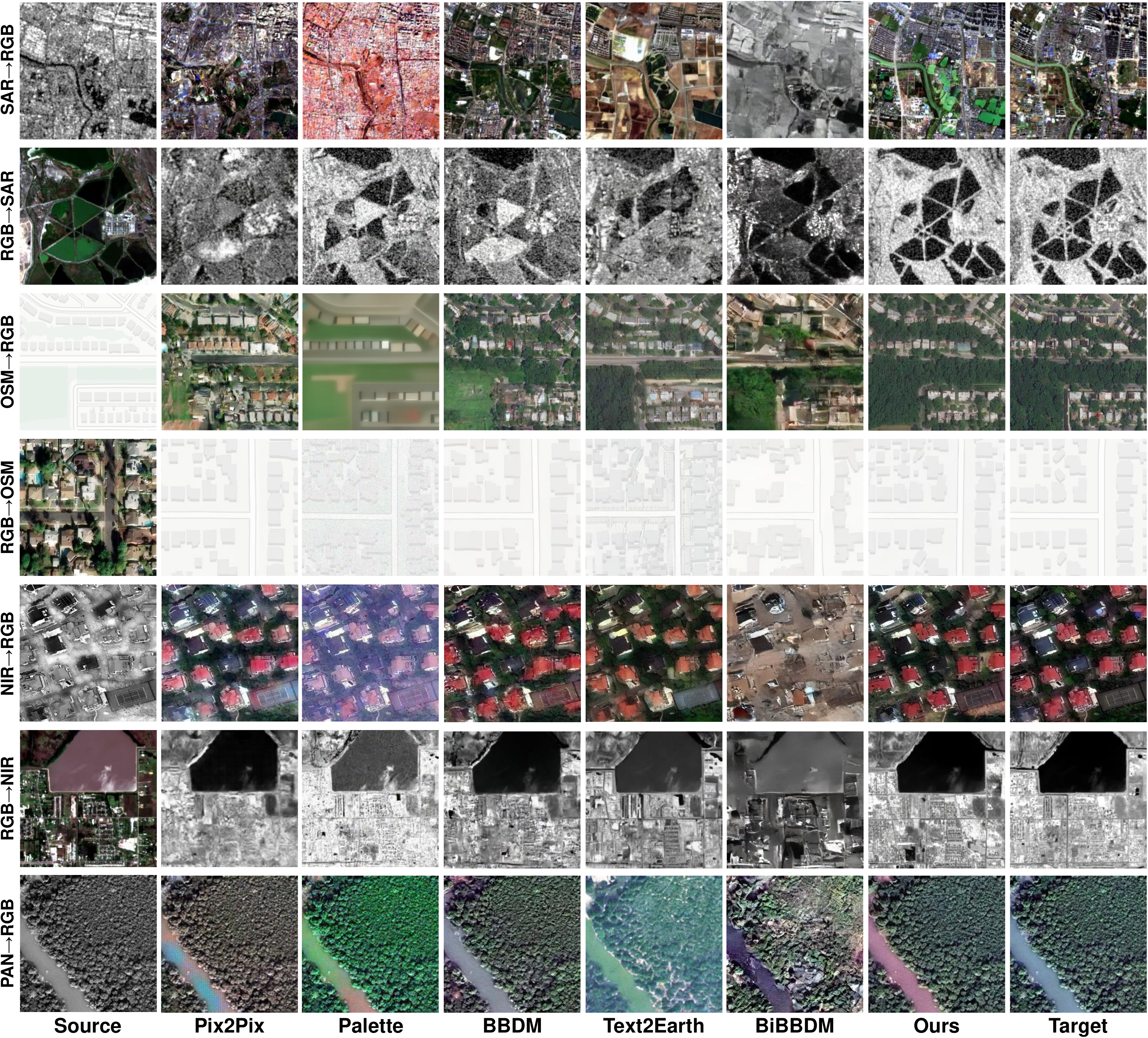}
    \caption{\textbf{Qualitative comparison on cross-modal translation.} MetaEarth-MM generates more realistic and cross-modally consistent images compared with representative image-to-image translation and diffusion-based methods across SAR/RGB, OSM/RGB, NIR/RGB, and PAN/RGB translation tasks.}
    \label{fig:exp_translation}
\end{figure*}

Tables~\ref{tab:exp_sar_osm_rgb} and \ref{tab:exp_nir_pan_rgb} report the quantitative results for cross-modal translation, and Fig.~\ref{fig:exp_translation} presents the corresponding qualitative comparisons. MetaEarth-MM achieves consistently strong performance across different modality groups and maintains a better balance between target-domain realism and source-condition consistency. These results show that MetaEarth-MM can handle different types of cross-modal gaps, including different imaging mechanisms, spectral responses, and levels of semantic abstraction.

\begin{table*}[t]
\centering
\setlength{\tabcolsep}{3pt}
\renewcommand{\arraystretch}{1.05}
\caption{Quantitative comparison of MetaEarth-MM and previous methods on SAR/RGB and OSM/RGB cross-modal translation.}
\label{tab:exp_sar_osm_rgb}
\begin{adjustbox}{max width=\textwidth}
\begin{tabular}{@{} c cccc cccc cccc cccc @{}}
\toprule
\multirow{2}{*}{Method} & \multicolumn{4}{c}{SAR$\rightarrow$RGB} & \multicolumn{4}{c}{RGB$\rightarrow$SAR} & \multicolumn{4}{c}{OSM$\rightarrow$RGB} & \multicolumn{4}{c}{RGB$\rightarrow$OSM} \\
\cmidrule(lr){2-5} \cmidrule(lr){6-9} \cmidrule(lr){10-13} \cmidrule(lr){14-17}
& FID$\downarrow$ & LPIPS$\downarrow$ & SSIM$\uparrow$ & CAS$\uparrow$
& FID$\downarrow$ & FSIM$\uparrow$ & PSNR$\uparrow$ & CAS$\uparrow$
& FID$\downarrow$ & LPIPS$\downarrow$ & SSIM$\uparrow$ & CAS$\uparrow$
& FID$\downarrow$ & LPIPS$\downarrow$ & SSIM$\uparrow$ & CAS$\uparrow$ \\
\midrule
Pix2Pix 
& 109.23 & 0.5864 & \textbf{0.2487} & 52.85 
& 138.21 & 0.6239 & 13.1573 & 45.11 
& 103.52 & 0.6253 & 0.2080 & 63.88 
& 37.80 & 0.2914 & \underline{0.8361} & \underline{62.18} \\
Palette 
& 144.81 & 0.6542 & 0.1231 & 49.17 
& 133.57 & 0.6582 & 9.6141 & 50.45 
& 168.74 & 0.7229 & 0.2062 & 54.25 
& 265.62 & 0.6443 & 0.5621 & 41.23 \\
BBDM 
& 91.18 & 0.5717 & 0.2349 & \underline{57.40} 
& 131.40 & \underline{0.6769} & \underline{13.1963} & \underline{56.09} 
& 86.80 & 0.5317 & \underline{0.2265} & \underline{67.90} 
& \underline{28.37} & \underline{0.2748} & 0.8191 & 60.94 \\
Text2Earth 
& \underline{49.43} & \underline{0.4943} & 0.2010 & 56.51 
& \underline{87.41} & 0.6601 & 11.8304 & 44.73 
& \underline{50.28} & \underline{0.5261} & 0.1908 & 64.83 
& 44.67 & 0.3543 & 0.7230 & 55.84 \\
BiBBDM 
& 182.26 & 0.6154 & 0.2293 & 42.62 
& 140.62 & 0.6302 & 10.6268 & 47.67 
& 214.66 & 0.5630 & 0.1744 & 52.96 
& 39.23 & 0.2958 & 0.8128 & 55.47 \\
\midrule
MetaEarth-MM 
& \multirow{2}{*}{\textbf{38.11}} & \multirow{2}{*}{\textbf{0.4392}} & \multirow{2}{*}{\underline{0.2441}} & \multirow{2}{*}{\textbf{59.82}} 
& \multirow{2}{*}{\textbf{47.10}} & \multirow{2}{*}{\textbf{0.6784}} & \multirow{2}{*}{\textbf{13.3415}} & \multirow{2}{*}{\textbf{60.04}} 
& \multirow{2}{*}{\textbf{43.04}} & \multirow{2}{*}{\textbf{0.5106}} & \multirow{2}{*}{\textbf{0.2415}} & \multirow{2}{*}{\textbf{70.04}} 
& \multirow{2}{*}{\textbf{23.08}} & \multirow{2}{*}{\textbf{0.2101}} & \multirow{2}{*}{\textbf{0.8411}} & \multirow{2}{*}{\textbf{68.17}} \\
(ours) & & & & & & & & & & & & & & & & \\
\bottomrule
\end{tabular}
\end{adjustbox}
\end{table*}

\begin{table*}[t]
\centering
\setlength{\tabcolsep}{5pt}
\renewcommand{\arraystretch}{1.05}
\caption{Quantitative comparison of MetaEarth-MM and previous methods on NIR/RGB and PAN/RGB cross-modal translation.}
\label{tab:exp_nir_pan_rgb}
\begin{adjustbox}{max width=\textwidth}
\begin{tabular}{@{}cccccccccccc@{}}
\toprule
\multirow{2}{*}{Method} 
& \multicolumn{4}{c}{NIR$\rightarrow$RGB} 
& \multicolumn{4}{c}{RGB$\rightarrow$NIR} 
& \multicolumn{3}{c}{PAN$\rightarrow$RGB} \\
\cmidrule(lr){2-5} \cmidrule(lr){6-9} \cmidrule(lr){10-12}
& FID$\downarrow$ & LPIPS$\downarrow$ & SSIM$\uparrow$ & CAS$\uparrow$
& FID$\downarrow$ & LPIPS$\downarrow$ & SSIM$\uparrow$ & CAS$\uparrow$
& FID$\downarrow$ & LPIPS$\downarrow$ & SSIM$\uparrow$ \\
\midrule
Pix2Pix
& \underline{30.47} & \underline{0.4332} & \textbf{0.5449} & \underline{67.54}
& \underline{35.93} & \textbf{0.1846} & \textbf{0.6477} & 66.73
& \underline{42.80} & \textbf{0.3205} & \textbf{0.8572} \\
Palette
& 87.66 & 0.4447 & 0.3821 & 60.01
& 133.70 & 0.3414 & 0.4448 & 62.77
& 143.61 & 0.3958 & 0.5925 \\
BBDM
& 68.91 & 0.4634 & 0.3754 & 66.76
& 76.32 & 0.3718 & 0.3946 & \underline{66.92}
& 51.60 & 0.3690 & 0.5661 \\
Text2Earth
& 41.55 & 0.4746 & 0.3095 & 61.60
& 50.29 & 0.3839 & 0.3782 & 63.82
& 51.28 & 0.3947 & 0.3331 \\
BiBBDM
& 134.52 & 0.5564 & 0.2906 & 60.52
& 118.41 & 0.5079 & 0.3045 & 54.57
& 128.74 & 0.4061 & 0.3148 \\
\midrule
MetaEarth-MM (ours)
& \textbf{28.45} & \textbf{0.4188} & \underline{0.3933} & \textbf{67.59}
& \textbf{32.61} & \underline{0.2940} & \underline{0.5199} & \textbf{68.38}
& \textbf{30.58} & \underline{0.3225} & \underline{0.7622} \\
\bottomrule
\end{tabular}
\end{adjustbox}
\end{table*}

For SAR/RGB translation, MetaEarth-MM shows clear advantages in both directions. In SAR$\rightarrow$RGB, it achieves an FID of 38.11 and the highest CAS, outperforming Text2Earth (FID: 49.43). This performance margin widens in RGB$\rightarrow$SAR. These results highlight the limitations of modality-specific priors. Previous methods heavily based on RGB pre-training often experience performance drops when the target domain shifts to SAR due to the large distribution gap. In contrast, MetaEarth-MM employs scene-centered modeling without any modality bias, which allows it to adapt to SAR-specific or RGB-specific appearance effectively. The qualitative results also show that MetaEarth-MM produces more coherent RGB textures and more faithful SAR structures, with fewer color shifts and over-smoothed regions.

For OSM/RGB translation, MetaEarth-MM shows stable performance despite the semantic abstraction gap between the two modalities. Specifically, OSM$\rightarrow$RGB demands synthesizing dense optical textures from sparse layouts, while RGB$\rightarrow$OSM requires extracting abstract structural representations from complex imagery. Across both directions, MetaEarth-MM consistently achieves the best FID, SSIM, and CAS. These results demonstrate that our method improves both appearance generation and structural abstraction. Qualitatively, MetaEarth-MM better preserves road topology and building layouts in OSM, and produces higher-fidelity roads and building blocks in the RGB imagery.

For NIR/RGB and PAN/RGB translation, the source and target modalities share stronger structural and appearance similarity. Although Pix2Pix obtains the highest SSIM in these tasks, as shown in Fig.~\ref{fig:exp_translation}, its outputs often suffer from severe blurring and noticeable artifacts. SSIM typically reward deterministic pixel-level mappings rather but fail to reflect perceptual realism. For example, in PAN$\rightarrow$RGB, Pix2Pix yields an FID of 42.80 despite its high SSIM, whereas MetaEarth-MM achieves a significantly lower FID of 30.58. Furthermore, our model obtains the best FID and CAS across all NIR/RGB directions. These results indicate that MetaEarth-MM effectively maintains strict cross-modal consistency and avoids the over-smoothed appearance common in previous methods.

Overall, MetaEarth-MM achieves superior and robust performance across diverse asymmetric translation tasks. Notably, BiBBDM consistently exhibits poor performance, ranking at the bottom across almost all tasks. This reveals that simply using one bidirectional model is insufficient for heterogeneous remote sensing translation, since opposite directions may involve conflicting generation and abstraction processes. By decoupling these processes through a latent scene representation, MetaEarth-MM avoids this optimization conflict, ensuring stable performance regardless of the translation direction.

\subsubsection{Comparison on unconditional paired joint generation}
We conduct comprehensive experiments on paired generation, focusing on OSM-RGB, SAR-RGB, and NIR-RGB tasks. Since most previous methods cannot generate aligned modality pairs directly, we construct a cascade pipeline to enable a fair comparison. In this setup, the first modality (e.g., OSM) is initially generated unconditionally from noise via MetaEarth-MM. The generated image is then mapped to the second modality (e.g., RGB) using various translation models. Since all cascade pipelines share identical first-stage output, their FID scores for the first modality are exactly the same. Any differences in the second-modality FID and cross-modal CAS strictly reflect the capability of the second-stage translation model.

Table \ref{tab:joint_generation} reports the quantitative results of these tasks. In the OSM-RGB generation task, the cascade pipeline generates OSM first and subsequently generates RGB. When evaluated and compared under the cascade pipeline, previous methods such as Pix2Pix, Palette, BBDM, Text2Earth, and BiBBDM exhibit clear limitations, producing lower-quality paired images and weaker cross-modal consistency. In contrast, utilizing MetaEarth-MM as the second-stage translator yields significantly better visual quality and alignment. This confirms its robustness: even when conditioned on imperfect generated samples rather than real observations, MetaEarth-MM consistently maintains high generation quality.

More importantly, our joint generation paradigm achieves the best overall performance, outperforming even the strong MetaEarth-MM cascade pipeline. In the OSM-RGB task, the joint generation obtains a noticeably lower RGB FID and a higher CAS than its cascaded pipeline. This result highlights that the paired generation is not simply equivalent to generating one modality and then translating it into another. In a cascade pipeline, generation errors from the first stage inevitably accumulate and mislead the second stage. Joint generation allows both modalities to mutually guide each other during the generation process. By jointly modeling their distribution, our model ensures that shared structural features are developed simultaneously, leading to more realistic individual distributions and tighter pairwise alignment. 

Considering that the cascade pipeline heavily relies on the second-stage translation, where previous methods have already shown worse performance, we only compare the joint and cascade generation of MetaEarth-MM on the SAR-RGB and NIR-RGB tasks. As demonstrated in the SAR-RGB and NIR-RGB sections of Table~\ref{tab:joint_generation}, joint generation consistently outperforms the cascade strategy, confirming that joint sampling is superior for paired multi-modal generation.

\begin{table*}[t]
\centering
\setlength{\tabcolsep}{3.5pt}
\renewcommand{\arraystretch}{1.08}
\caption{Quantitative comparison on unconditional paired generation. For OSM-RGB, cascade methods first generate OSM using MetaEarth-MM and then translate it into RGB using different translation models. Starred values (*) denote the shared first-stage OSM results generated by MetaEarth-MM. For SAR-RGB and NIR-RGB, MetaEarth-MM Cascade refers to the MetaEarth-MM cascade pipeline.}
\label{tab:joint_generation}
\resizebox{\textwidth}{!}{%
\begin{tabular}{cccccccc|ccc|ccc}
\toprule
\multicolumn{8}{c|}{OSM-RGB}
& \multicolumn{3}{c|}{SAR-RGB}
& \multicolumn{3}{c}{NIR-RGB} \\
\cmidrule(lr){1-8} \cmidrule(lr){9-11} \cmidrule(lr){12-14}
\multirow{2}{*}{Metric}
& \multirow{2}{*}{Joint}
& \multicolumn{6}{c|}{Cascade method}
& \multirow{2}{*}{Metric}
& \multirow{2}{*}{Joint}
& \multirow{2}{*}{\shortstack{MetaEarth-MM\\Cascade}}
& \multirow{2}{*}{Metric}
& \multirow{2}{*}{Joint}
& \multirow{2}{*}{\shortstack{MetaEarth-MM\\Cascade}} \\
\cmidrule(lr){3-8}
& & Pix2Pix & Palette & BBDM & Text2Earth & BiBBDM & MetaEarth-MM
& & & 
& & & \\
\midrule
FID$_{\rm OSM}$$\downarrow$
& \textbf{16.92}
& 17.36* & 17.36* & 17.36* & 17.36* & 17.36* & 17.36
& FID$_{\rm SAR}$$\downarrow$
& \textbf{61.36} & 62.08
& FID$_{\rm NIR}$$\downarrow$
& \textbf{44.24} & 46.24 \\
FID$_{\rm RGB}$$\downarrow$
& \textbf{45.68}
& 115.98 & 167.11 & 70.04 & 59.40 & 207.89 & 46.86
& FID$_{\rm RGB}$$\downarrow$
& \textbf{30.47} & 34.85
& FID$_{\rm RGB}$$\downarrow$
& \textbf{38.11} & 39.92 \\
CAS$\uparrow$
& \textbf{64.45}
& 57.86 & 58.38 & 62.06 & 57.92 & 45.59 & 62.42
& CAS$\uparrow$
& \textbf{52.76} & 50.90
& CAS$\uparrow$
& \textbf{63.60} & 61.57 \\
\bottomrule
\end{tabular}%
}
\end{table*}

\subsection{Ablation study}

\begin{table*}[t]
\centering
\scriptsize
\setlength{\tabcolsep}{2.4pt}
\renewcommand{\arraystretch}{1.08}
\caption{Ablation study of MetaEarth-MM. The contributions of the decoupled architecture, modality-routed FFN, and scene consistency constraint are evaluated across cross-modal translation and paired joint generation tasks.}
\label{tab:exp_ablation}
\begin{adjustbox}{max width=\textwidth}
\begin{tabular}{@{}ccc|ccc|ccc|ccc|ccc|ccc@{}}
\toprule
\multirow{2}{*}{\shortstack{Decoupled\\ architecture}}
& \multirow{2}{*}{\shortstack{Modality-routed\\FFN}}
& \multirow{2}{*}{\shortstack{Scene consistency\\constraint}}
& \multicolumn{3}{c|}{SAR$\rightarrow$ RGB}
& \multicolumn{3}{c|}{OSM$\rightarrow$ RGB}
& \multicolumn{3}{c|}{RGB$\rightarrow$ NIR}
& \multicolumn{3}{c|}{SAR-RGB}
& \multicolumn{3}{c}{SAR-NIR} \\
\cmidrule(lr){4-6}
\cmidrule(lr){7-9}
\cmidrule(lr){10-12}
\cmidrule(lr){13-15}
\cmidrule(lr){16-18}
& & 
& FID$\downarrow$ & LPIPS$\downarrow$ & CAS$\uparrow$
& FID$\downarrow$ & LPIPS$\downarrow$ & CAS$\uparrow$
& FID$\downarrow$ & LPIPS$\downarrow$ & CAS$\uparrow$
& FID$_{\rm RGB}\downarrow$ & FID$_{\rm SAR}\downarrow$ & CAS$\uparrow$
& FID$_{\rm NIR}\downarrow$ & FID$_{\rm SAR}\downarrow$ & CAS$\uparrow$ \\
\midrule
$\times$ & -- & --
& 54.18 & 0.4784 & 55.08
& 58.39 & 0.5294 & 65.94
& 41.59 & 0.3728 & 64.37
& 39.77 & 65.94 & 49.12
& 50.56 & 70.67 & 48.68 \\

$\checkmark$ & $\times$ & $\times$
& 48.78 & 0.4682 & 55.75
& 52.11 & 0.5199 & 66.49
& 35.61 & 0.3454 & 66.44
& 37.49 & 64.43 & 50.80
& 47.41 & 71.98 & 48.93 \\

$\checkmark$ & $\checkmark$ & $\times$
& 44.54 & 0.4609 & 56.47
& 49.45 & 0.5170 & 66.90
& 34.99 & 0.3448 & 66.61
& 36.14 & 64.26 & 49.92
& 45.62 & 71.33 & 49.17 \\

$\checkmark$ & $\checkmark$ & $\checkmark$
& \textbf{39.93} & \textbf{0.4437} & \textbf{59.38}
& \textbf{45.82} & \textbf{0.5146} & \textbf{69.46}
& \textbf{33.74} & \textbf{0.3417} & \textbf{66.71}
& \textbf{33.81} & \textbf{63.13} & \textbf{51.05}
& \textbf{44.76} & \textbf{68.84} & \textbf{50.26} \\
\bottomrule
\end{tabular}
\end{adjustbox}

\vspace{1mm}
\footnotesize
\end{table*}

\begin{figure*}[t]
    \centering
    \includegraphics[width=\textwidth]{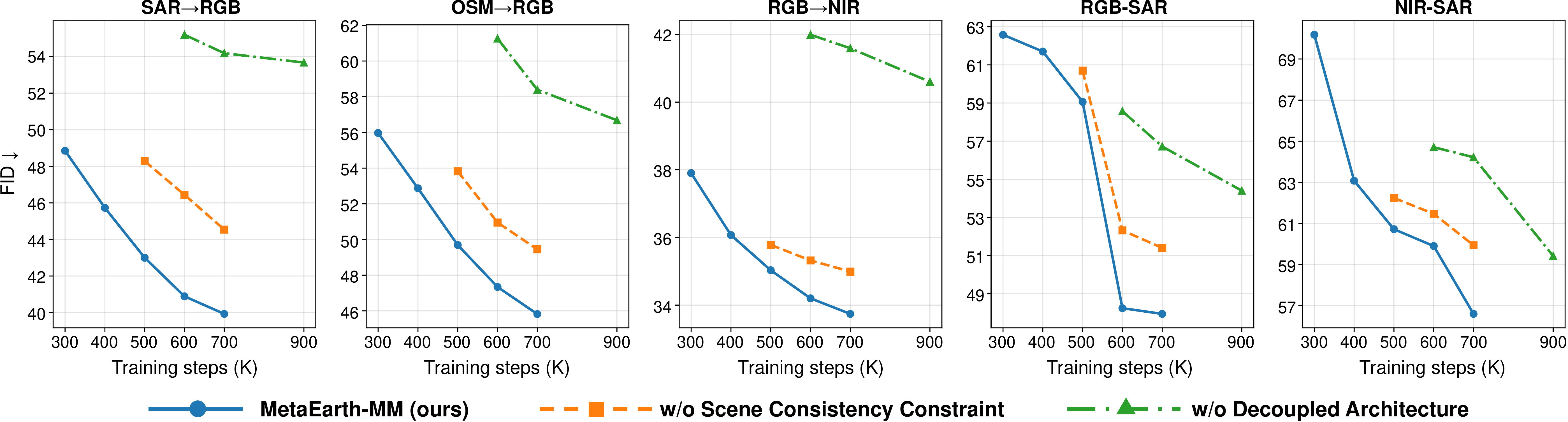}
    \caption{FID evaluation curves during training for MetaEarth-MM and its ablated variants on cross-modal translation and paired joint generation tasks. The x-axis denotes training steps, and the y-axis denotes test-set FID. For paired joint generation tasks, the reported FID is computed as the average FID of the two generated modalities. MetaEarth-MM achieves faster convergence and better generation quality.}
    \label{fig:exp_ablation}
\end{figure*}

We conduct ablation studies to evaluate the core designs of MetaEarth-MM: the scene-centered decoupled architecture, the modality-routed FFN, and the scene consistency constraint. Table \ref{tab:exp_ablation} compares the full MetaEarth-MM against progressively degraded variants. The baseline is a direct joint transformer, which uses a standard ViT with an equivalent total depth but lacks explicit decoupling between scene inference and modality generation. All models are trained for 700k steps under identical setups.

Overall, the full MetaEarth-MM achieves the best FID, LPIPS, and CAS across all conditional translation and paired generation tasks. More importantly, the progressive settings reveal the specific contributions of each component:

\textbf{Decoupled architecture.} Comparing the decoupled baseline with the direct joint transformer reveals that architectural decoupling alone significantly improves performance. In a standard joint transformer, cross-modal information aggregation and modality-specific flow prediction are heavily entangled. Our design introduces an intermediate representation as an explicit interface, providing a separate space to reorganize information before generation. This advantage is particularly pronounced in pairs with massive domain gaps (e.g. SAR/RGB and OSM/RGB), where modeling direct appearance-level correspondence is highly challenging.

\textbf{Modality-routed FFN.} Introducing the modality-routed FFN further reduces FID. Forcing a fully shared generator to fit heterogeneous texture and structural distributions of different modalities inevitably causes cross-modality competition. The routed FFN mitigates this by assigning dedicated parameter capacity to each modality, thereby enhancing individual generation quality.

\textbf{Scene Consistency Constraint.} This constraint imposes a scene-level prior on the intermediate representation, encouraging the scene inference module to organize it around the underlying scene content. With this prior, the decoupled architecture achieves a clearer role assignment: scene inference module provides a scene-level basis, while the generator focuses exclusively on representing the scene in the modality-specific manner and dedicates its computational capacity entirely to modeling the characteristics of the target modality, resulting in improved cross-modal correspondence and generation fidelity.

\textbf{Optimization Stability.} Finally, the training curves (Fig. \ref{fig:exp_ablation}) show that MetaEarth-MM converges faster and reaches a lower final FID. Under the same training budget, the direct joint transformer and other ablated settings plateau at higher FID scores. These results indicate that the proposed method not only improves the final generation quality but also makes the multi-modal generation objective easier to optimize.

\section{Downstream Applications}
Beyond image generation, we further examine how MetaEarth-MM can support downstream tasks at data-level and representation-level. This provides a functional evaluation of whether our model captures transferable cross-modal knowledge during training, highlighting its potential as a versatile foundation model for multi-modal Earth observation.

\subsection{Generative data augmentation}
Paired multi-modal data are essential for many remote sensing tasks, such as cross-modal matching, image registration, and data fusion, where consistent observations of the same scene across different modalities are required. However, such data are often expensive to collect at scale and difficult to align accurately. To address this data scarcity, we serve MetaEarth-MM as a generative data engine for data augmentation, generating high-quality, aligned modality pairs to expand training sets. In this section, we evaluate this capability on cross-modal image matching, a representative task that directly tests whether the generated pairs preserve useful spatial and semantic correspondence.

We consider two cross-modal matching tasks, RGB-OSM and RGB-SAR. For each task, we start from 10K real paired training samples and use MetaEarth-MM to generate an additional 10K paired samples, resulting in an augmented training set of 20K samples. The generated pairs are used only for training augmentation, while evaluation is conducted on real paired test samples. Ground-truth correspondences are obtained from known random geometric transformations. To cover different matching paradigms, we evaluate three representative methods: LightGlue~\cite{lightglue} for sparse matching, LoFTR~\cite{loftr} for semi-dense matching, and DKM~\cite{dkm} for dense matching. Performance is measured using EPE and PCK under different pixel thresholds.

\begin{table*}[t]
\centering
\setlength{\tabcolsep}{4pt}
\renewcommand{\arraystretch}{1.08}
\caption{Results of generative data augmentation for cross-modal image matching. For each matcher, we compare models trained with real paired data only and with additional generated pairs. The best result between the two training settings is shown in bold.}
\label{tab:exp_downstream_matching}
\resizebox{\textwidth}{!}{
\begin{tabular}{llcccccccccc}
\toprule
\multirow{2}{*}{Matcher} 
& \multirow{2}{*}{Training Data}
& \multicolumn{5}{c}{RGB-OSM}
& \multicolumn{5}{c}{RGB-SAR} \\
\cmidrule(lr){3-7} \cmidrule(lr){8-12}
& 
& EPE$\downarrow$ & PCK@1px$\uparrow$ & PCK@3px$\uparrow$ & PCK@5px$\uparrow$ & PCK@10px$\uparrow$
& EPE$\downarrow$ & PCK@1px$\uparrow$ & PCK@3px$\uparrow$ & PCK@5px$\uparrow$ & PCK@10px$\uparrow$ \\
\midrule

\multirow{2}{*}{LightGlue}
& Real 
& 3.757 & 0.07798 & 0.4824 & 0.7804 & 0.9647
& 3.560 & 0.09955 & 0.5697 & 0.8480 & \textbf{0.9730} \\
& Real+Gen 
& \textbf{3.494} & \textbf{0.0959} & \textbf{0.5416} & \textbf{0.8116} & \textbf{0.9695}
& \textbf{3.489} & \textbf{0.1047} & \textbf{0.5948} & \textbf{0.8595} & 0.9720 \\
\midrule

\multirow{2}{*}{LoFTR}
& Real 
& 5.574 & 0.1711 & 0.3841 & 0.5830 & 0.8823
& 20.16 & 0.2102 & 0.3464 & 0.4935 & 0.7831 \\
& Real+Gen 
& \textbf{4.900} & \textbf{0.2480} & \textbf{0.4432} & \textbf{0.6357} & \textbf{0.9005}
& \textbf{4.932} & \textbf{0.3344} & \textbf{0.4845} & \textbf{0.6451} & \textbf{0.8828} \\
\midrule

\multirow{2}{*}{DKM}
& Real 
& 28.44 & 0.006399 & 0.05235 & 0.1182 & 0.2770
& 39.83 & 0.004074 & 0.03343 & 0.07826 & 0.1974 \\
& Real+Gen 
& \textbf{11.29} & \textbf{0.02295} & \textbf{0.1705} & \textbf{0.3397} & \textbf{0.6213}
& \textbf{10.27} & \textbf{0.02352} & \textbf{0.1730} & \textbf{0.3521} & \textbf{0.6647} \\

\bottomrule
\end{tabular}
}
\end{table*}

As shown in Table~\ref{tab:exp_downstream_matching}, adding jointly generated pairs consistently improves matching performance across both modality pairs and all three matchers. The gains are particularly pronounced for LoFTR and DKM, suggesting that the generated pairs provide not only greater scene diversity but also strong pixel-level alignment. The improvement is even larger on RGB-SAR, where the modality gap is much larger than that of RGB-OSM. This further indicates that MetaEarth-MM can generate paired samples with stable and accurate cross-modal correspondence even when the visual differences between modalities are substantial.

\subsection{Image-level domain adaptation}

Modern visual perception models are often pretrained on large-scale RGB imagery, which makes the RGB domain a major source of reusable visual knowledge. However, remote sensing modalities such as SAR and NIR have visual appearances that differ greatly from RGB images, making it difficult to directly apply this knowledge to non-RGB data. A common solution is to adapt the model itself by initializing from RGB-pretrained weights and fine-tuning on the target modality. In this section, we conduct an exploratory experiment from a data-centric perspective: before downstream fine-tuning, we use MetaEarth-MM to translate non-RGB observations into the RGB domain, bringing the input data closer to the domain where pretrained visual knowledge is available.

We evaluate this data-centric adaptation setting on semantic segmentation using the WHU-OPT-SAR~\cite{whu} dataset. Given SAR or NIR inputs, MetaEarth-MM first translates them into RGB images. The segmentation models are initialized with RGB-pretrained weights and then fine-tuned on the translated training images, followed by evaluation on the translated test images. The pretrained MetaEarth-MM is used directly, without any task-specific fine-tuning. As a comparison, we fine-tune the same RGB-pretrained segmentation models directly on the original SAR, and NIR inputs. To examine whether the effect is consistent across architectures, we adopt two representative segmentation models: the transformer-based MaskFormer~\cite{maskformer} and the CNN-based DeepLabv3~\cite{deeplabv3}.

\begin{table}[t]
\centering
\footnotesize
\setlength{\tabcolsep}{3.5pt}
\renewcommand{\arraystretch}{1.08}
\caption{Semantic segmentation results on WHU-OPT-SAR. SAR$\rightarrow$RGB and NIR$\rightarrow$RGB denote translated RGB images. For each source modality, the better result between the original input and translated input is shown in bold.}
\label{tab:exp_downstream_segmentation}
\begin{tabular}{cc|cc|cc}
\toprule
Method & Metric 
& SAR 
& SAR$\rightarrow$RGB 
& NIR 
& NIR$\rightarrow$RGB \\
\midrule
\multirow{2}{*}{MaskFormer}
& mIoU$\uparrow$ 
& 14.53 & \textbf{38.53} & 36.83 & \textbf{43.28} \\
& mAcc$\uparrow$ 
& 27.96 & \textbf{51.22} & 47.40 & \textbf{55.00} \\
\midrule
\multirow{2}{*}{DeepLabv3}
& mIoU$\uparrow$ 
& 28.49 & \textbf{33.58} & 35.17 & \textbf{39.04} \\
& mAcc$\uparrow$ 
& 37.46 & \textbf{43.45} & 46.99 & \textbf{50.92} \\
\bottomrule
\end{tabular}
\end{table}

As shown in Table~\ref{tab:exp_downstream_segmentation}, translating heterogeneous modalities into the RGB domain leads to substantial improvements in segmentation performance. The effect is especially pronounced for SAR, where direct SAR inputs perform poorly for both MaskFormer and DeepLabv3, yielding only 14.53 and 28.49 mIoU, respectively. After SAR$\rightarrow$RGB translation, the performance rises sharply to 38.53 and 33.58 mIoU. NIR$\rightarrow$RGB translation also consistently outperforms direct NIR input on both models. These results validate the effectiveness of this data-level adaptation strategy for heterogeneous remote sensing modalities. By translating SAR and NIR inputs into the RGB domain, MetaEarth-MM places the inputs in a domain where RGB-pretrained visual priors can be more effectively activated. The consistent improvements further suggest that the translated images are not only visually plausible, but also preserve semantic information useful for downstream tasks.

\subsection{Zero-shot representation transfer}
Cross-modal understanding is a key challenge in remote sensing, since recognition models trained on one modality often suffer severe degradation when applied to another directly. Beyond using MetaEarth-MM for generative data augmentation and image-level domain adaptation, we further examine whether its internal scene representation itself supports zero-shot representation transfer. Specifically, we probe the latent scene tokens inferred by the scene inference module and evaluate them on scene classification without retraining the model.

We construct a seven-class scene classification dataset from SEN12-MS~\cite{sen12ms}, including urban, cropland, forest, grassland, shrubland, savanna, and bare land. Each class contains 600 training samples and 200 test samples, with aligned RGB, SAR, and NIR images. This setting allows us to train a classifier based on one modality and test it on another modality.

We freeze the scene inference module $\mathcal{E}_\theta$ and train only a lightweight attention-pooling classifier $\mathcal{C}_\phi$. Given a training modality $m_i$ and another modality $m_j$, the classifier is trained on the scene tokens inferred from the clean latent of $m_i$:
\begin{equation}
[\hat{\mathbf{s}}_i, \hat{\mathbf{s}}_j]
=
\mathcal{E}_\theta
\left(
\mathbf{z}_i^{0}, \mathbf{z}_j^{1}, 0, 1, m_i, m_j
\right),
\qquad
\hat{y}=\mathcal{C}_\phi(\hat{\mathbf{s}}_i).
\end{equation}
During zero-shot inference, $m_i$ is unavailable and its branch is replaced with random noise, while the clean latent of $m_j$ is provided to the other branch:
\begin{equation}
[\hat{\mathbf{s}}_{i|j}, \hat{\mathbf{s}}_j]
=
\mathcal{E}_\theta
\left(
\mathbf{z}_i^{1}, \mathbf{z}_j^{0}, 1, 0, m_i, m_j
\right),
\qquad
\hat{y}=\mathcal{C}_\phi(\hat{\mathbf{s}}_{i|j}).
\end{equation}
Here, $\hat{\mathbf{s}}_{i|j}$ denotes the $m_i$-branch scene tokens inferred from modality $m_j$. The classifier is kept frozen, so the evaluation directly measures whether the inferred scene representation is transferable across modalities.

\begin{table}[t]
\centering
\setlength{\tabcolsep}{10pt}
\renewcommand{\arraystretch}{1.0}
\caption{Zero-shot modality transfer results on scene classification task. Accuracy (\%) is reported. In-domain results are shown in gray, and the best result in each train-test setting is shown in bold.}
\label{tab:exp_zero_shot}
\begin{tabular}{ll ccc}
\toprule
\multicolumn{2}{c}{Modality Setting} & \multicolumn{3}{c}{Method} \\
\cmidrule(lr){1-2} \cmidrule(lr){3-5}
Train & Test & ResNet50 & DINOv3 & MetaEarth-MM \\
\midrule
\multirow{3}{*}{RGB} 
& {\color{gray}RGB} & {\color{gray}83.57} & {\color{gray}\textbf{89.07}} & {\color{gray}85.29} \\
& SAR               & 18.57 & 18.21 & \textbf{65.00} \\
& NIR               & 44.50 & 66.57 & \textbf{76.14} \\
\midrule
\multirow{2}{*}{SAR} 
& {\color{gray}SAR} & {\color{gray}68.64} & {\color{gray}\textbf{75.93}} & {\color{gray}75.14} \\
& RGB               & 19.86 & 25.50 & \textbf{68.29} \\
\midrule
\multirow{2}{*}{NIR} 
& {\color{gray}NIR} & {\color{gray}76.79} & {\color{gray}\textbf{88.14}} & {\color{gray}82.36} \\
& SAR               & 14.71 & 20.57 & \textbf{65.00} \\
\bottomrule
\end{tabular}
\end{table}

As shown in Table~\ref{tab:exp_zero_shot}, MetaEarth-MM is compared with ResNet50~\cite{resnet} and DINOv3~\cite{dinov3} under the same frozen-backbone setting, where only a lightweight classifier is trained. ResNet50 is pretrained on our constructed dataset, and DINOv3 uses its official remote-sensing version. MetaEarth-MM achieves consistently higher accuracy across all zero-shot train-test modality pairs, while ResNet50 and DINOv3 degrade substantially once the test modality differs from the training modality. On average, MetaEarth-MM preserves 83.82\% of the accuracy obtained on the training modality, compared with only 30.89\% for ResNet50 and 38.03\% for DINOv3. These results indicate that the latent scene representation learned by MetaEarth-MM captures modality-transferable scene semantics, making it a promising modality-agnostic feature extractor for zero-shot downstream tasks.

\section{Conclusion}
In this paper, we propose MetaEarth-MM, a unified generative foundation model for multi-modal remote sensing imagery. Instead of learning isolated conditional cross-modal mappings, MetaEarth-MM introduces a scene-centered joint modeling paradigm that organizes multi-modal generation around intrinsic scene content. Built upon a decoupled architecture comprising a scene inference module and a modality-aware routed generator, the model supports paired joint generation and any-to-any translation across five modalities. We also construct EarthMM, a large-scale multi-modal remote sensing dataset with 2.8 million globally distributed images, providing a data foundation for training and evaluation. Experiments show that MetaEarth-MM achieves consistently superior generation quality and cross-modal alignment across diverse translation and joint generation tasks, while further enabling zero-shot generation on unseen modality combinations. Beyond serving as a powerful multi-modal generative data engine, MetaEarth-MM learns transferable scene-level semantics, supporting downstream tasks through generative data augmentation, image-level domain adaptation, and zero-shot representation transfer. These results demonstrate its potential as a foundation model for cross-modal Earth observation.

\bibliography{content}

\begin{IEEEbiography}[{\includegraphics[width=1in,height=1.25in,clip,keepaspectratio]{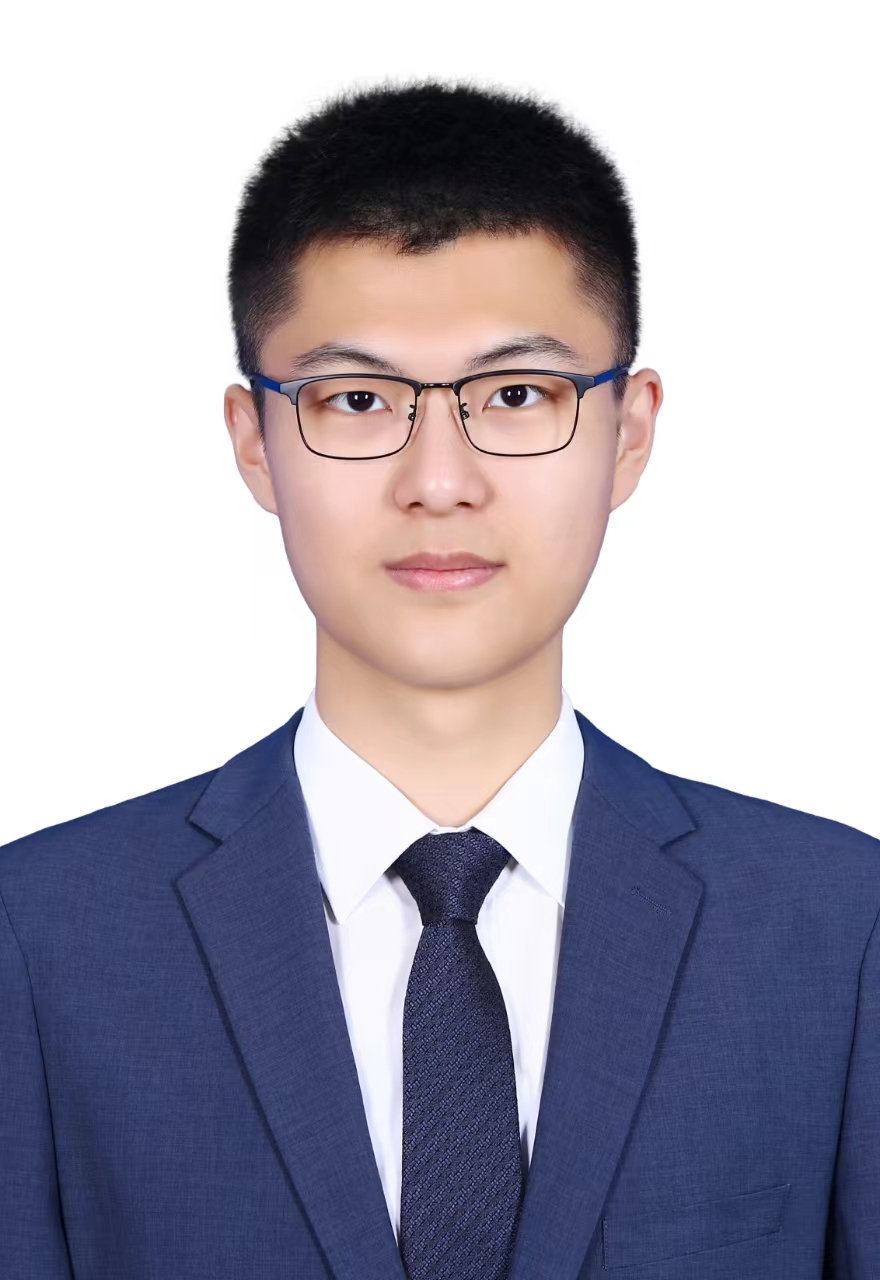}}]{Zhiping Yu} received his B.S. degree from the Department of Aerospace Intelligent Science and Technology, School of Astronautics, Beihang University in 2025. He is currently working towards the Ph.D. degree in the Department of Aerospace Intelligent Science and Technology, School of Astronautics, Beihang University.
His research interests include deep learning and remote sensing image generation.
\end{IEEEbiography}

\begin{IEEEbiography}[{\includegraphics[width=1in,height=1.25in,clip,keepaspectratio]{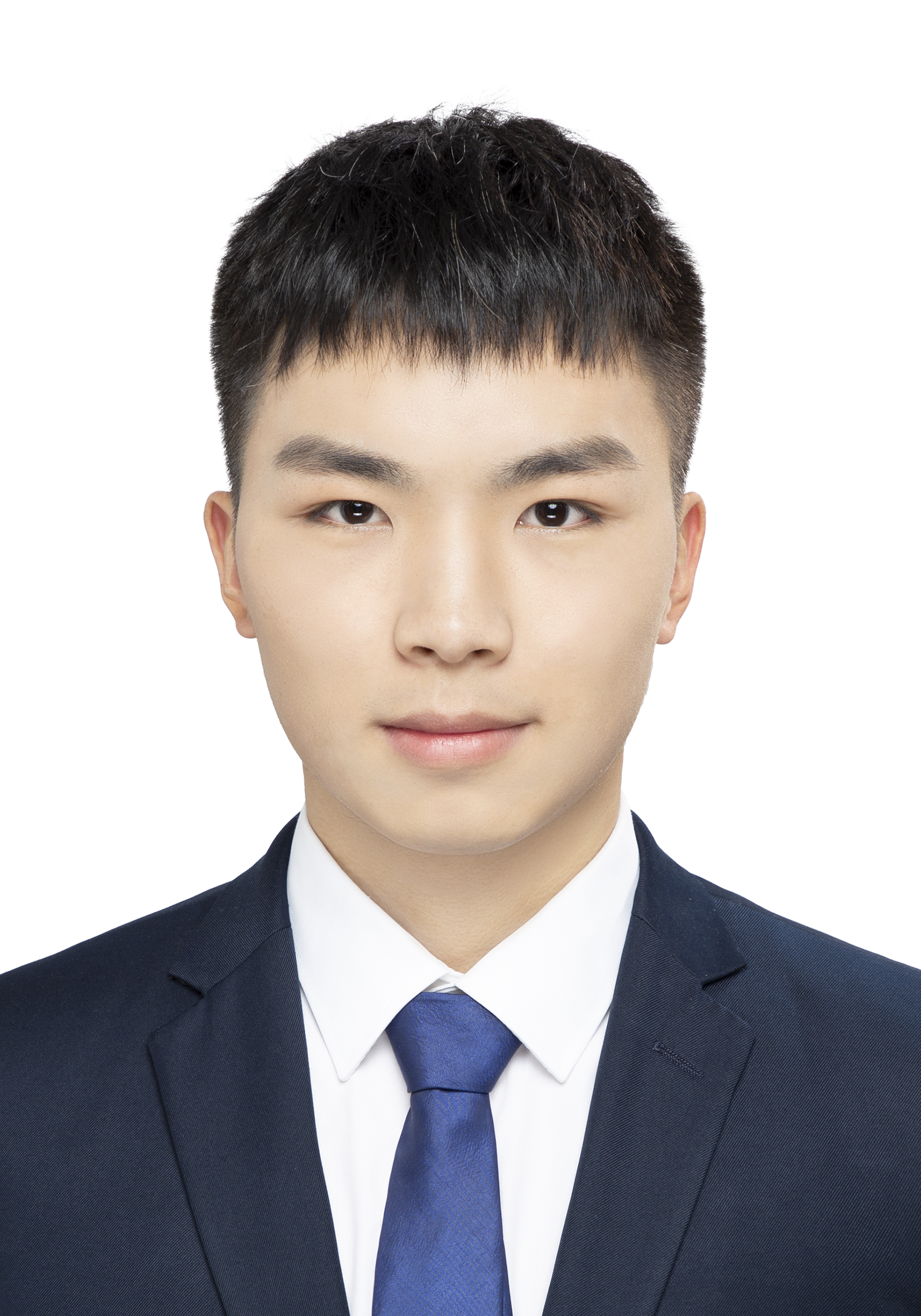}}]{Chenyang Liu} received his B.S. degree from the Image Processing Center, School of Astronautics, Beihang University in 2021. He is currently working towards the Ph.D. degree in the Department of Aerospace Intelligent Science and Technology, School of Astronautics, Beihang University. His research interests include machine learning, computer vision, and multimodal learning.
\end{IEEEbiography}

\begin{IEEEbiography}[{\includegraphics[width=1in,height=1.25in,clip,keepaspectratio]{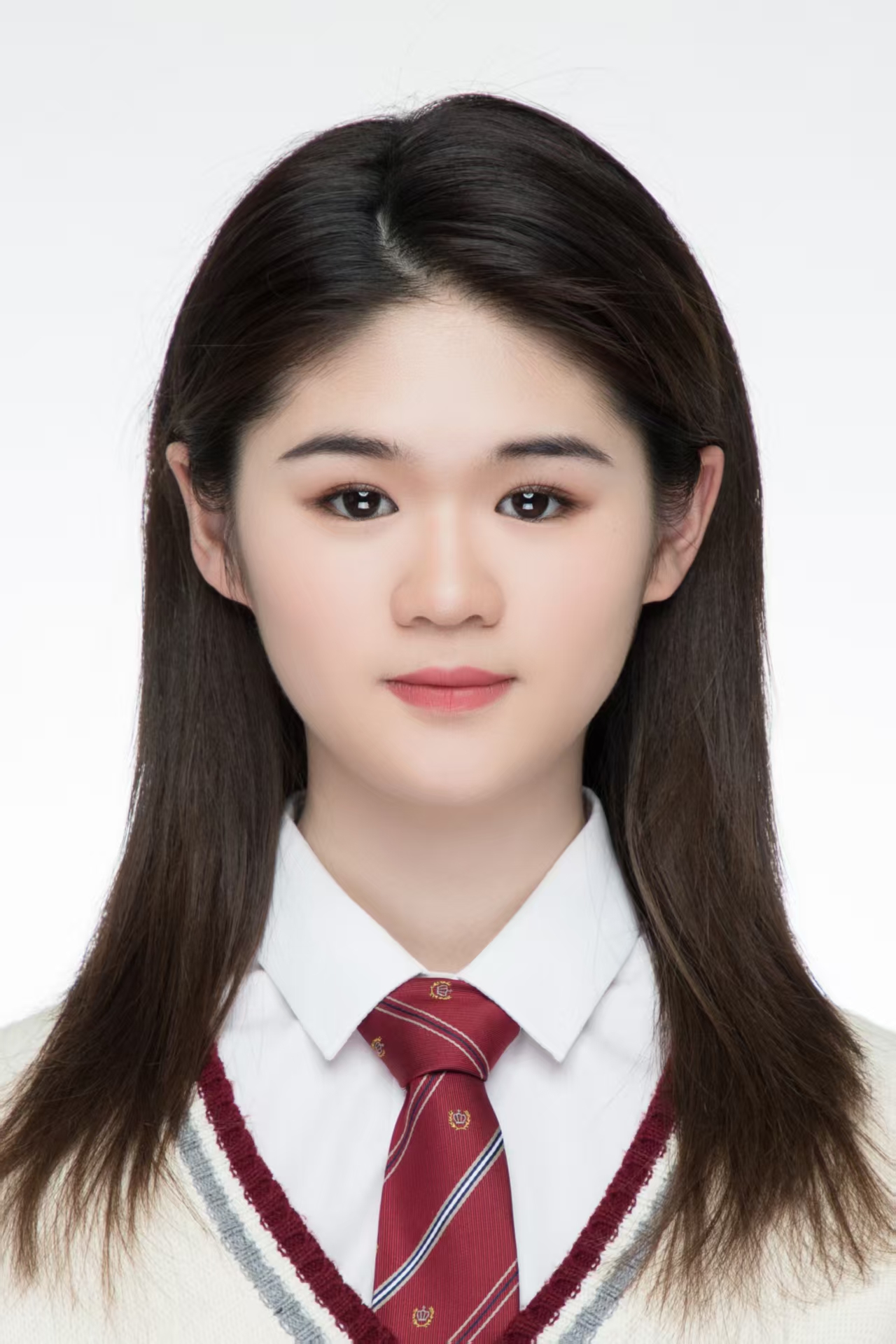}}]{Jinqi Cao} is currently an undergraduate student at the Department of Aerospace Intelligent Science and Technology, School of Astronautics, Beihang University. Her research interests include deep learning and remote sensing image processing.
\end{IEEEbiography}

\begin{IEEEbiography}[{\includegraphics[width=1in,height=1.25in,clip,keepaspectratio]{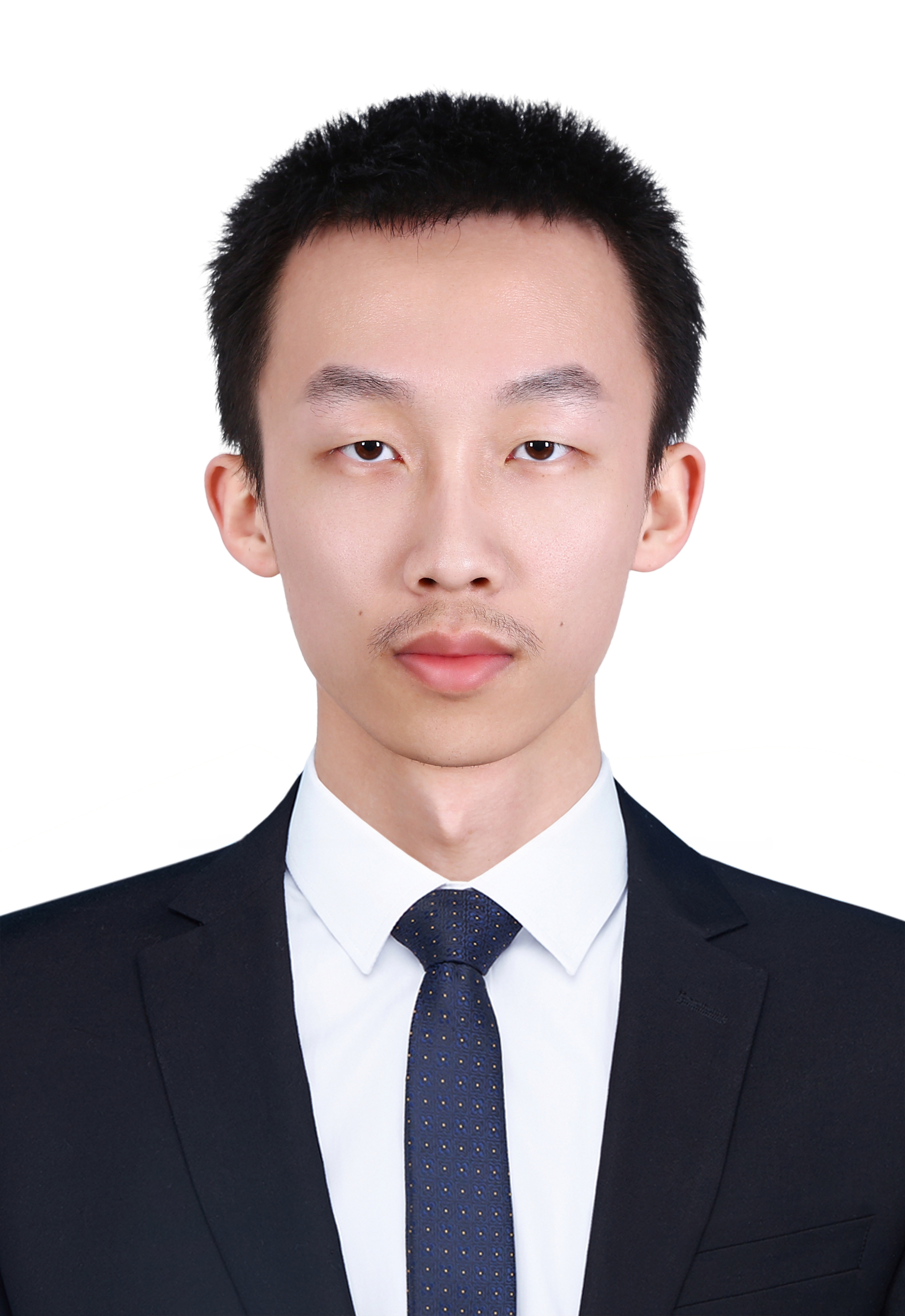}}]{Qinzhe Yang} is currently a student at the Shen Yuan Honors College, Beihang University, pursuing a Ph.D. degree through an eight-year integrated program in Future Aerospace Technology. His research interests include image processing and deep learning, particularly object detection and instance segmentation in remote sensing.
\end{IEEEbiography}

\begin{IEEEbiography}[{\includegraphics[width=1in,height=1.25in,clip,keepaspectratio]{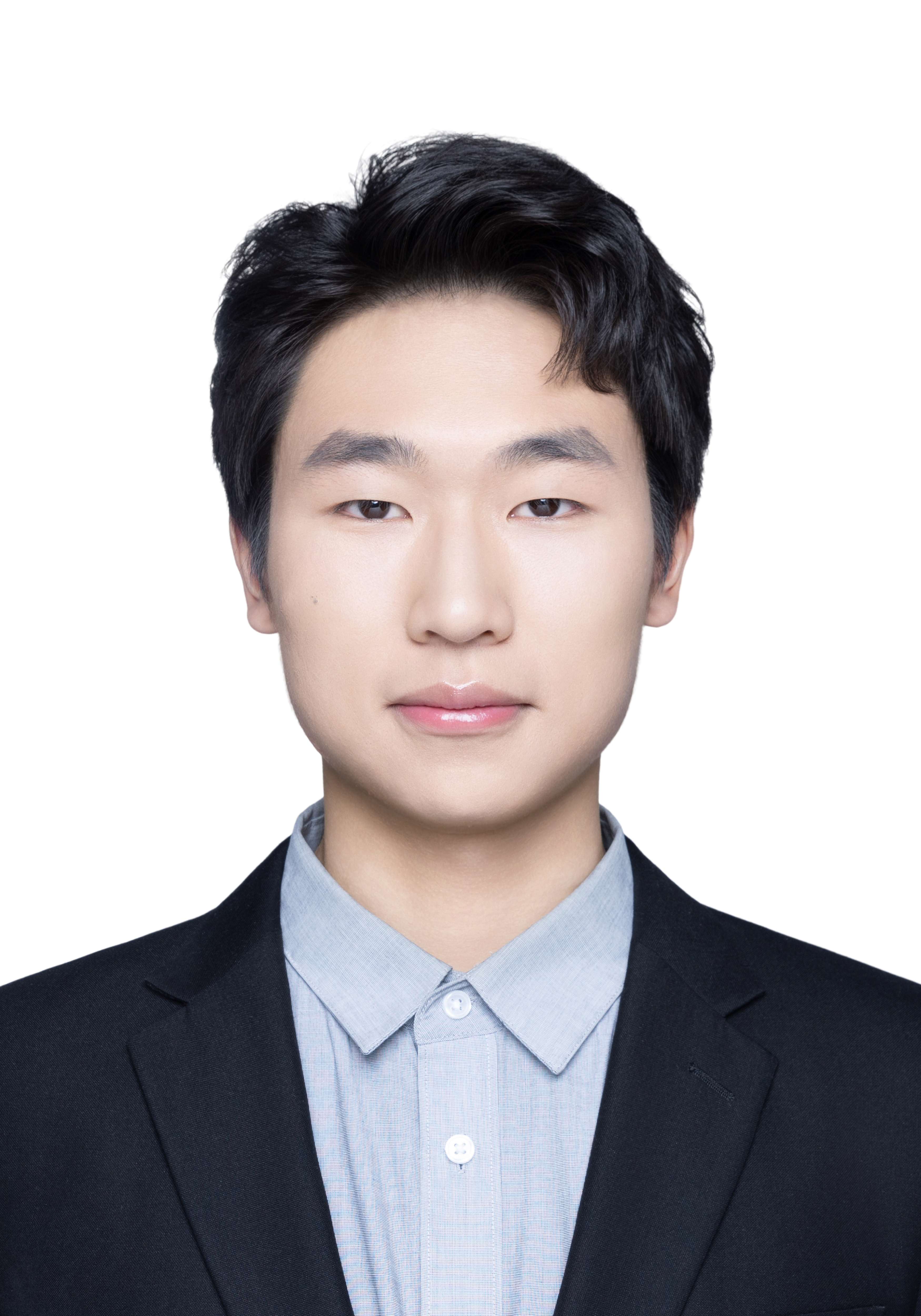}}]{Siwei Yu} received his B.S. degree from the Department of Aerospace Intelligent Science and Technology, School of Astronautics, Beihang University in 2025. He is currently working towards the Ph.D. degree in the Department of Aerospace Intelligent Science and Technology, School of Astronautics, Beihang University.
His research interests include deep learning and remote sensing image registration.
\end{IEEEbiography}

\begin{IEEEbiography}[{\includegraphics[width=1in,height=1.25in,clip,keepaspectratio]{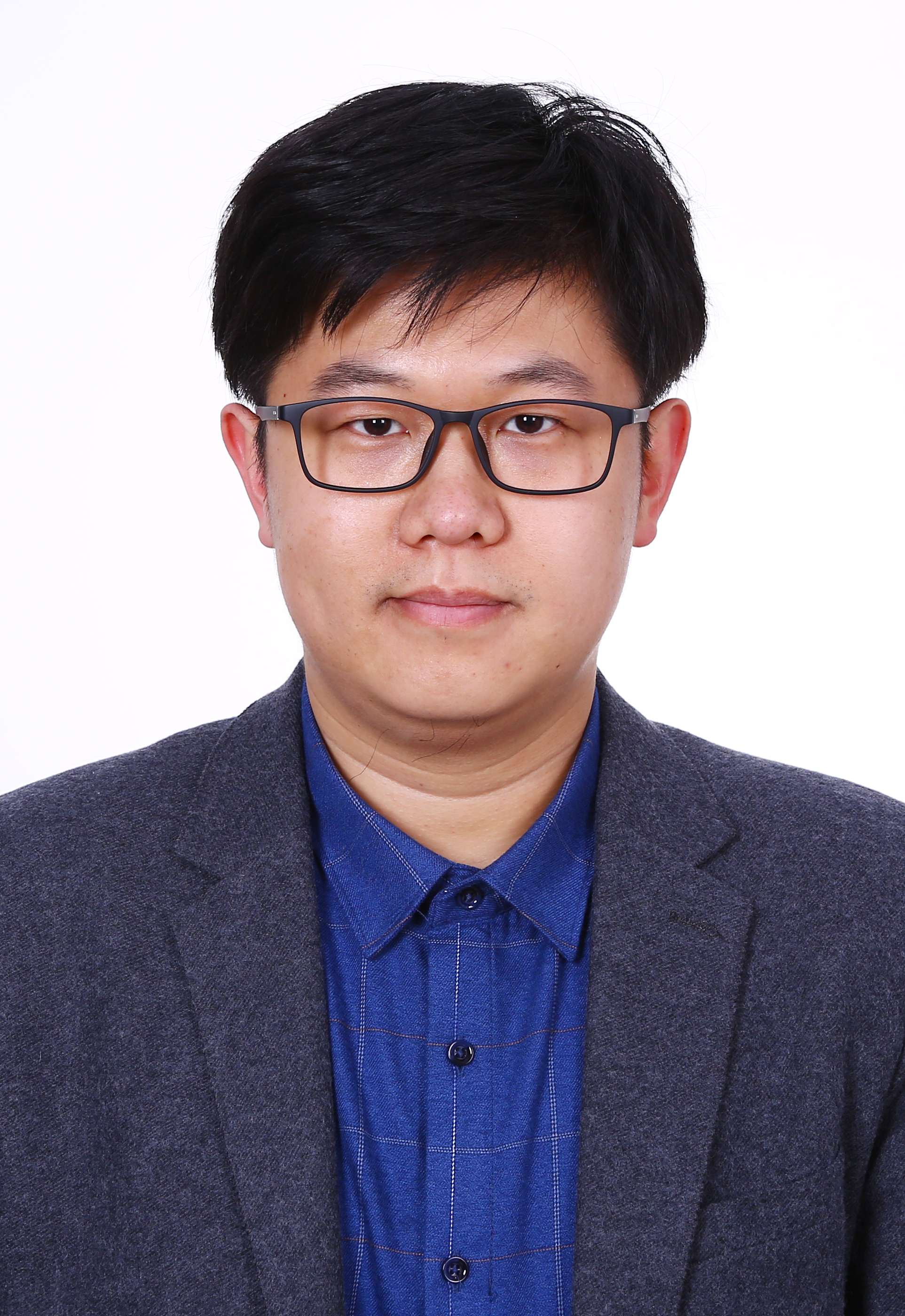}}]{Zhengxia Zou}(Senior Member, IEEE) received his BS degree and his Ph.D. degree from Beihang University in 2013 and 2018. He is currently a Professor at the Department of Aerospace Intelligent Science and Technology, School of Astronautics, Beihang University. During 2018-2021, he was a postdoc research fellow at the University of Michigan, Ann Arbor. His research interests include computer vision and related problems in remote sensing. He has published over 100 peer-reviewed papers in top-tier journals and conferences, including Proceedings of the IEEE, Nature Communications, IEEE Transactions on Pattern Analysis and Machine Intelligence, IEEE Transactions on Geoscience and Remote Sensing, and IEEE / CVF Computer Vision and Pattern Recognition. 
 
Dr. Zou serves as the Associate Editor for IEEE Transactions on Image Processing and IEEE Transactions on Geoscience and Remote Sensing. His personal website is \url{https://zhengxiazou.github.io/}.
\end{IEEEbiography}

\begin{IEEEbiography}[{\includegraphics[width=1in,height=1.25in,clip,keepaspectratio]{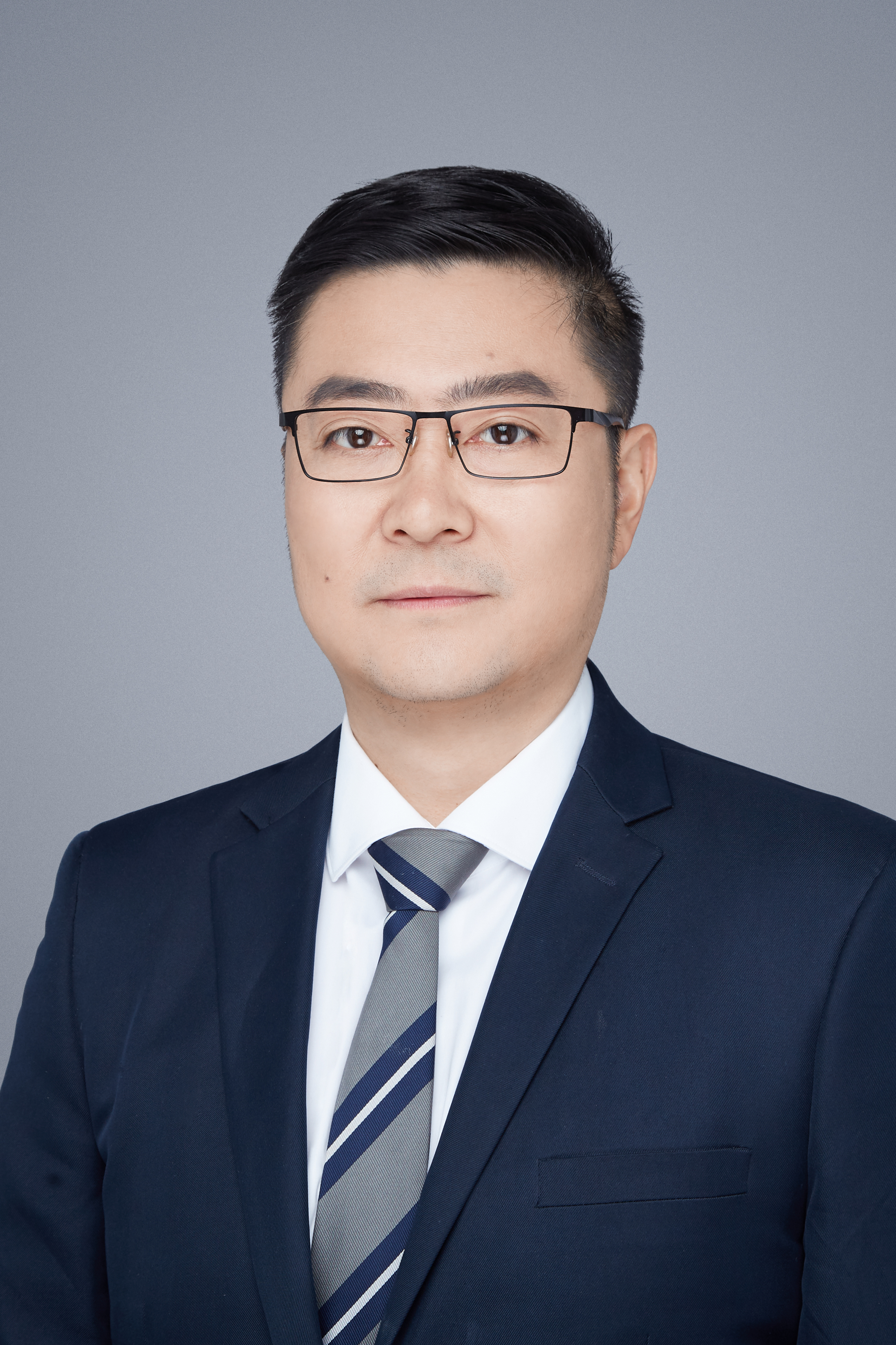}}]{Zhenwei Shi}(Senior Member, IEEE) is currently a Professor and Dean of the Department of Aerospace Intelligent Science and Technology, School of Astronautics, Beihang University. He has authored or co-authored over 300 scientific articles in refereed journals and proceedings, including the IEEE Transactions on Pattern Analysis and Machine Intelligence, the IEEE Transactions on Image Processing, the IEEE Transactions on Geoscience and Remote Sensing, the IEEE Conference on Computer Vision and Pattern Recognition (CVPR) and the IEEE International Conference on Computer Vision (ICCV). His current research interests include remote sensing image processing and analysis, computer vision, pattern recognition, and machine learning.

Dr. Shi serves as an Editor for the IEEE Transactions on Geoscience and Remote Sensing, the Pattern Recognition, the Remote Sensing, and the Infrared Physics and Technology, etc. 
His personal website is \url{http://levir.buaa.edu.cn/}..
\end{IEEEbiography}

\end{document}